\documentclass[12pt,a4paper]{article}

\usepackage[british]{babel}

\usepackage[a4paper,top=2cm,bottom=2cm,left=2.5cm,right=2.5cm,marginparwidth=1.75cm]{geometry}
\usepackage{lineno}
\usepackage{natbib}
\usepackage{hyperref}
\usepackage[dvipsnames]{xcolor}
\usepackage{amsmath}
\usepackage{graphicx}
\usepackage[title]{appendix}
\usepackage{mathrsfs}
\usepackage{amsfonts}
\usepackage{booktabs} 
\usepackage{caption}  
\usepackage{threeparttable} 
\usepackage{algorithm}
\usepackage{algorithmicx}
\usepackage{algpseudocode}
\usepackage{listings}
\usepackage{enumitem}
\usepackage{chngcntr}
\usepackage{booktabs}
\usepackage{lipsum}
\usepackage{subcaption}
\usepackage{authblk}
\usepackage[T1]{fontenc}    
\usepackage{csquotes}       
\usepackage{diagbox}
\usepackage{babel}
\usepackage{multirow}
\usepackage{array}
\usepackage{caption}
\usepackage{graphicx}

\newcommand{\btheta}{\boldsymbol{\theta}}
\newcommand{\bx}{\boldsymbol{x}}

\newcommand{\by}{\boldsymbol{y}}

\newcommand{\argmin}{\operatorname*{argmin}}
\newcommand{\rev}{\textcolor{blue}}

\usepackage{setspace}
\usepackage{titlesec}
\titleformat{\section} 
  {\normalfont\Large\bfseries}{\thesection.}{1em}{}
  
\usepackage{float}   
\usepackage{caption} 


\begin{document}

\title{Testing and Improving the Robustness of \\ Amortized Bayesian Inference for Cognitive Models}
\date{}
\author[1,*]{Yufei Wu}
\author[2]{Stefan T. Radev}
\author[1]{Francis Tuerlinckx}

\affil[1]{University of Leuven, Belgium}
\affil[2]{Rensselaer Polytechnic Institute, US}

\affil[*]{\textit{Corresponding author:}yufei.wu@kuleuven.be}

\maketitle
\begin{abstract}
\noindent Contaminant observations often cause problems when estimating the parameters of cognitive models. In this study, we tested and improved the robustness of parameter estimation using amortized Bayesian inference. We conducted systematic analyses in two settings: a toy example (i.e., a normal distribution with an unknown mean) and a popular cognitive model, the Drift Diffusion Model. First, we studied the stylized sensitivity curve and the breakdown point of the estimators. Next, we proposed a simple data augmentation approach that incorporated a contamination distribution into the data-generating process during training to train robust estimators. We examined several robust estimators with different contamination distributions, and evaluated their performance and cost in terms of accuracy and efficiency loss relative to a standard estimator. Introducing contaminants from a Cauchy distribution during training significantly increases the robustness of the neural density estimator, as measured by bounded sensitivity functions and a substantially higher breakdown point. Overall, the proposed method is straightforward and practical to implement, with broad applicability in fields where outlier detection or removal is challenging. 
\end{abstract}



\section*{Introduction}
Identifying and appropriately handling contaminants is both unavoidable and essential for drawing accurate conclusions from quantitative data. Contaminants are generated by processes different from the process under study. For example, in behavioral science, careless responding is a major source of contamination \citep{maniaci_caring_2014}. On the one hand, contamination can lead to outliers or extreme values that deviate significantly from the majority of data points \citep[]{aggarwal_outlier_2017,hawkins_identification_1980}.
Commencing with Huber's foundational work (\citeyear{huber_robust_1964}) on M-estimators for estimating the mean of a univariate normal distribution, the field of robust statistics has generated several methods for detecting and dealing with outliers \citep[]{donoho_notion_1982, hawkins_identification_1980, tukey_robust_1979, hampel_robust_2005}. On the other hand, contaminants can also have numerical values that are difficult to distinguish from the majority, making accurate estimates more challenging \citep{maronna_robust_2006}.

Some statistical and computational models in psychology are particularly vulnerable to the presence of contaminants due to the nature of their underlying assumptions. One example is the widely used Drift Diffusion Model \citep[DDM;][]{ratcliff_theory_1978} that explains patterns in choice reaction time data from fast decision-making tasks. In the most basic DDM variant, a key parameter is the non-decision time ($T_{er}$) capturing the time taken for stimulus encoding and motor execution. By design, $T_{er}$ is estimated to be lower than the shortest reaction time in the data set. Thus, a short outlier can distort the estimate of $T_{er}$ and consequently the other parameters that jointly determine the process with $T_{er}$. As a result, mitigating the impact of contaminants has been a persistent challenge in DDM estimation \citep[]{ratcliff_methods_1993, ratcliff_modeling_1998, ratcliff_estimating_2002,myers_practical_2022}.

Various sophisticated methods have been proposed to reduce the impact of outliers in fitting of different models \citep[]{maronna_robust_2006,sumarni_robustness_2017,aggarwal_outlier_2017}.
When models are fit using maximum likelihood estimation (MLE) or Bayesian methods \citep[e.g., MCMC sampling,][]{kruschke_doing_2011}, robust methods often require modifying the loss or the likelihood function of the model. This introduces two key challenges. First, modifying the loss function can be impractical for many model users. Second, not all models have an explicitly available loss function or likelihood \citep{palestro2018likelihood}.

In this paper, we studied robustness within an amortized Bayesian inference \citep[ABI,][]{radev_bayesflow_2020} framework, a simulation-based approach combined with neural networks. ABI leverages simulated training data to optimize generative neural networks that encode structural and functional knowledge about the simulation model and its parameters. By recasting the costly posterior sampling task as forward passes through a trained neural network, ABI achieves nearly instantaneous (i.e., \textit{amortized}) parameter inference for new data sets \citep[]{gonccalves2020training, radev_bayesflow_2020}. Throughout this paper, we refer to these neural networks as neural posterior estimators (NPEs). Crucially, ABI methods bypass the model likelihood (hence their alternative designation as ``likelihood-free'') and avoid expensive MCMC sampling or likelihood approximations altogether, making them an attractive and versatile addition to the modeler's toolkit \citep[]{schmitt2024amortized}.

Nevertheless, amortized methods are more susceptible to estimation errors than their non-amortized counterparts (e.g., MCMC) in the presence of \textit{model misspecification} \citep[]{schmitt_detecting_2021}.  The presence of contaminants can be viewed as a special case of model misspecification \citep[]{schmitt_detecting_2021}, and it is a serious obstacle for the adoption of amortized methods across behavioral and cognitive sciences, where computational models are nearly always approximations and data tends to be noisy, unpredictable, and corrupted in unexpected ways. Although previous work has focused on inducing \textit{robust} amortized estimators \citep[]{ward_robust_2022, huang2024learning, siahkoohi2023reliable, kelly2024misspecification, elsemuller2025does}, these methods either require non-trivial modifications to the straightforward simulation-based training or sacrifice the amortization property.

Thus, in this study, we investigated the influence of contaminants and proposed a simple data augmentation approach to enhance the robustness of ABI in cognitive modeling. To clarify the use of ABI, we compared amortized and traditional inference methods and conducted a systematic empirical investigation into the meaning of learned summary statistics. The univariate normal distribution and DDM served as test beds to demonstrate our robust methodology. Our approach can be extended to a wide range of other stochastic models of cognition, whether likelihood-based or simulation-based.

\subsection*{Amortized Bayesian Inference}

Amortized methods utilize model simulations to train specialized neural networks that learn to compress data of varying sizes and sample from the posterior of model parameters given \textit{any} data set compatible with the model \citep{burkner2023some}.

\begin{figure}[t]
\centering
\includegraphics[width=0.99\linewidth]{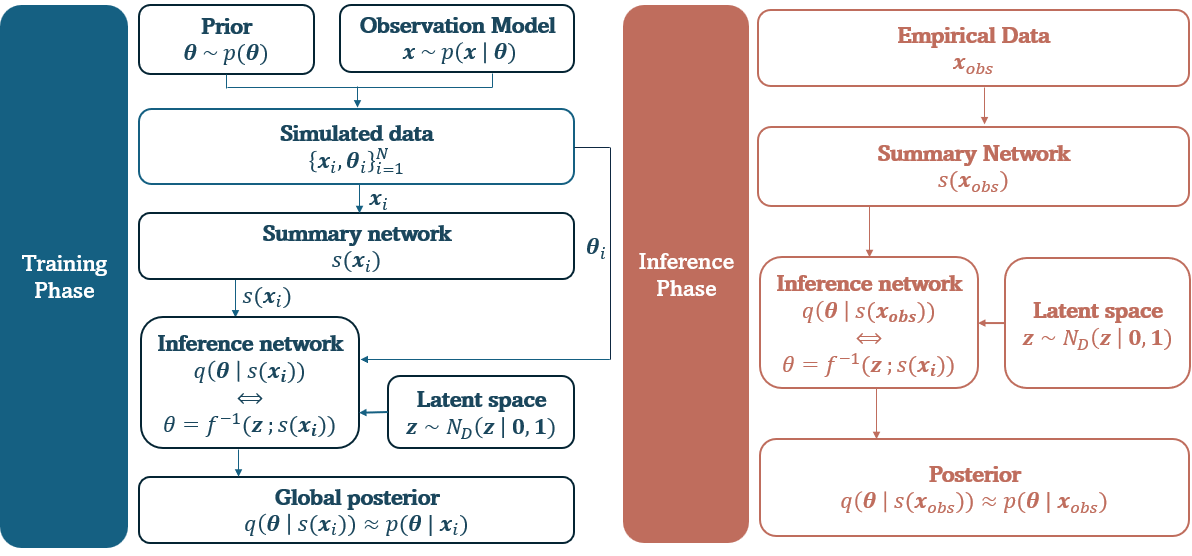}
\caption{\label{fig:BayesFlowworkflow}A basic amortized Bayesian workflow. Parameters and data are simulated from a prior and an observation model. The simulations are used as training data for the summary and inference networks that jointly learn the posterior. Once the networks are trained, they can instantly sample from the approximate posterior given any new data.}
\end{figure}

A simple amortized workflow is depicted in Figure~\ref{fig:BayesFlowworkflow}. During the training phase, a prior $\btheta \sim p(\btheta)$ and a generative model $\bx \sim p(\bx \mid \btheta)$ are defined, where $\btheta \in \mathbb{R}^D$. Subsequently, parameters are simulated from the prior and passed to the cognitive model to generate synthetic data sets. The simulated parameters and data sets $\{\bx_i, \btheta_i\}_{i=1}^I$ are used to train two neural networks: a summary and an inference network. The summary network transforms each data set $\bx_i$ into fixed-size \textit{approximately sufficient} summary statistics $s(\bx_i)$, where $s(\cdot)$ represents the transformation applied by the summary network. Simultaneously, the inference network learns to approximate the true posterior of model parameters given $\bx$ as accurately as possible:
\begin{equation}
    q(\btheta\,|\,s(\bx)) \approx p(\btheta\,|\,\bx),  
\end{equation}

Approximating complex posterior distributions involves sampling from high-dimensional conditional probability distributions. Recent advancements in generative deep learning provide a wide range of expressive algorithms to address this challenge \citep[e.g.,][]{kobyzev2020normalizing, lipmanflow, yang2023diffusion}.
For instance, normalizing flows \citep{kobyzev2020normalizing} learn an invertible transformation $f$ between the complex target distribution and a predefined simple latent distribution $\boldsymbol{z}$ (e.g., a spherical Gaussian), such that sampling from $\boldsymbol{z}$ and applying the inverse $f^{-1}$ yields samples from the approximate posterior:
\begin{equation}
    \btheta \sim q(\btheta \mid s(\bx)) \Longleftrightarrow \btheta = f^{-1} (\boldsymbol{z};\,s(\bx)) \mbox{ with } \boldsymbol{z} \sim \mathcal{N}_{D}(\boldsymbol{\boldsymbol{z}\,|\, 0}, \mathbf{I}), \label{eq:normalizing flow}
\end{equation}
where $f$ is an invertible function parameterized by a conditional invertible network \citep[]{ardizzone_guided_2019, radev_bayesflow_2020}.
The summary and inference networks are jointly optimized to ensure that the approximate posterior corresponds as closely as possible to the true posterior. 
The parameters of the neural networks are learned by minimizing the Kullback-Leibler (KL) divergence between the true and the approximate posterior for any data set $\bx$ sampled from the prior predictive distribution $p(\bx)$ \citep[for more details, please refer to][]{kullback_information_1951,radev_bayesflow_2020}:
\begin{equation}
    (f^*, s^*) = \argmin_{f, s}\mathbb{E}_{p(\bx)}\Bigl[D_\mathrm{KL}\bigl(p(\btheta \mid \bx)\,||\,q(\btheta \mid s(\bx)\bigr)\Bigr]. \label{eq:KL} 
\end{equation}
Upon convergence, the neural networks acquire an approximate representation of the full posterior. In practice, we define a prior distribution broad enough to generate a realistic range of empirical data. The inference phase (see Figure~\ref{fig:BayesFlowworkflow}) is then straightforward: the observed data $\bx^{(obs)}$ is fed into the networks, and the corresponding posterior can be inferred without any overhead.

More recent amortized algorithms employ multi-step continuous-time models, such as flow-matching \citep{dax2023flow}, diffusion models \citep{simons2023neural}, and few-step consistency models \citep{schmitt2024consistency}. In this study, we focused on normalizing flows, the simplest family of deep probabilistic models that works sufficiently well in practice.

\subsection*{Drift Diffusion Model}
The DDM is by far the most popular cognitive model for fitting choice reaction time data in terms of neurocognitively plausible parameters \citep[]{ratcliff_theory_1978, ratcliff_diffusion_2016,usher_time_2001}. In choice reaction tasks, participants are asked to make binary decisions, such as determining the moving direction (left or right) of a collection of random dots \citep[]{mulder_bias_2012}, or classifying words and non-words \citep[] {ratcliff_diffusion_2004}. Performance is then compared across different conditions (e.g., primed vs. non-primed) or participant demographics (e.g., younger vs. older adults). 

\begin{figure}
\centering
\includegraphics[width=1\linewidth]{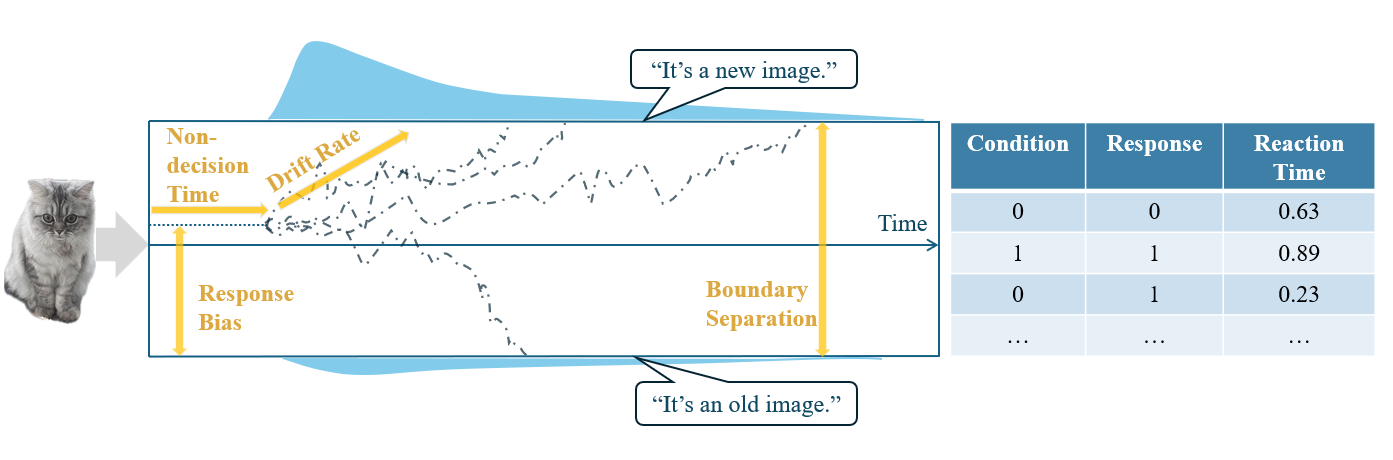}
\caption{\label{fig:DDMprocess}A graphical illustration of the drift diffusion process and the resulting reaction time data in a hypothetical visual recognition memory task. Participants view an image and judge whether it is old (i.e., previously seen) or new (i.e., previously unseen). The rows in the resulting data table correspond to individual trials of the experiment, with conditions and responses coded as 0 (old image) and 1 (new image).}
\end{figure}

The core assumptions of DDM are illustrated in Figure~\ref{fig:DDMprocess}. Participants accumulate evidence about a certain decision after stimulus onset, and when the evidence reaches one of the two boundaries: 0 or $a$, a decision is made, and $a$ is known as the boundary separation. The starting point of this process is denoted as $z \in [0,1]$. If $z>0.5$ ($z<0.5$), then there is a bias towards the upper (lower) boundary. The drift rate $v$ represents the average rate of evidence accumulation under a specific condition. The stochastic differential equation (SDE) provides a precise description of this diffusion process:

\begin{equation}
    dX_t = v \, dt + \sigma \, dW_t,
\label{eq:1dwiener}
\end{equation}

for $0<X_t<a$. Equation~\ref{eq:1dwiener} captures the random evolution of accumulated evidence $X$ over time $t$, with $v$, the deterministic drift, and $dW_t$, the differential of Brownian motion (or Wiener process). When $X_t \geq a$ or $X_t \leq 0$, the diffusion process stops and the decision is made. Finally, the non-decision time $T_{er}$ accounts for processes unrelated to decision formation. These parameters together determine the distribution of reaction times and choices.
 
Numerous variants of the standard DDM have been proposed \citep[see e.g.,][]{ratcliff_estimating_2002, ratcliff_modeling_1998,ratcliff_diffusion_2016,wieschen_jumping_2020}. This paper focuses on the standard version. The reasons for this choice are threefold. First, the added parameters, such as trial-to-trial variability parameters in $v$ and $T_{er}$, often cause estimation problems \citep[see e.g.,][]{lerche2016model,tillman2020sequential}. Second, the standard DDM is an interesting case to study robustness, given the strong dependence of $T_{er}$ on the minimum response time. Third, its analytical tractability enables us to better understand the mechanisms underlying learned summary statistics in ABI, effectively unpacking the learning process's black box.

Although DDM is known to be sensitive to outliers and various fitting methods exist for this model \citep{ratcliff_estimating_2002, wagenmakers_ez-diffusion_2007, voss_fast_2008, hBayesDM}, no study has systematically investigated the impact of contaminants in DDM estimation\citep{ratcliff_estimating_2002, lerche2017many}. One reason for this gap is that large-scale simulation studies can be highly time-consuming with existing methods, while the instantaneous posterior sampling with ABI is extremely efficient \citep[e.g.,][]{von_krause_mental_2022}. 

There is no universal solution to the problem of contamination. A common approach is to set cutoffs for reaction times (standard thresholds include reaction times longer than 100ms or 200ms and shorter than 2.5s or 3s) and discard trials that fall outside these predetermined limits \citep{myers_practical_2022}. Alternative methods to mitigate the impact of outliers include applying cutoffs based on either the standard deviation or the interquartile range \citep[]{ratcliff_methods_1993}. While being less arbitrary, these methods can reduce information gain and introduce their own biases \citep{ratcliff_methods_1993, ulrich_effects_1994}. More challenges arise when masked contaminants are present in data sets \citep{ratcliff_methods_1993}, making the DDM an ideal test bed for evaluating our data augmentation approach within the framework of ABI.

\section*{Parameter Recovery Study for the Drift Diffusion Model}

To offer insights into the performance and inner workings of end-to-end amortized inference, we compared estimates from (a) ABI, implemented in the Python package \texttt{BayesFlow} \citep[]{radev_bayesflow_2023}, version 1.1.4, (b) MCMC sampling, using the R library \texttt{JAGS} \citep{plummer2003jags} and its \texttt{dwiener} module \citep{wabersich_extending_2014}, and (c) EZ diffusion \citep{wagenmakers_ez-diffusion_2007}. As the golden standard for fully Bayesian estimation, MCMC sampling serves as a benchmark to evaluate ABI's performance. EZ diffusion helps to understand the learned summary statistics in ABI, because its estimation relies on three \textit{sufficient summary statistics}. It works only when the model assumes $z = 0.5$ and provides an analytical solution to the other key DDM parameters, $v$, $a$, and $T_{er}$. EZ diffusion uses the following sufficient summary statistics: $M_{RT}$, the mean reaction time of correct responses; $V_{RT}$, the variance of reaction time for correct responses; and $P_c$, the proportion of correct response \citep[for details, please refer to][]{wagenmakers_ez-diffusion_2007}. We estimated the DDM parameters with BayesFlow, JAGS, and the EZ-diffusion to investigate (a) the validity of our amortized Bayesian setup for recovering the parameters compared to simpler alternatives; and (b) the relationship between the summary statistics learned by the summary network and the simple statistics used by EZ-diffusion.

For simulation, we assumed a minimal DDM with $z = 0.5$ and a single drift rate $v$. Ground-truth parameters $(v, a, T_{er})$ were sampled from their corresponding prior (Table~\ref{priorpr}), with the within-trial noise variance $\sigma^2$ in the diffusion process fixed at 1. The prior was selected to align with the conservative parameter range suitable for EZ diffusion, based on a systematic review of DDM parameters. \citep{wagenmakers_ez-diffusion_2007, tran2021systematic}. Each data set contained 200 observations, and every observation $i$ (with $i=1,\dots,200$) consisted of a pair of a reaction time $x_i$ and choice $y_i$, jointly simulated from the Wiener process, $(x_i,y_i) \sim \mbox{Wiener}(v,a,z=0.5,T_{er})$.

For ABI, we defined the neural network architecture based on model and data requirements. Given the DDM's IID trial assumption, we used a permutation-invariant Set Transformer \citep{lee_set_2019} as the summary network, with three outputs matching the EZ diffusion method's sufficient statistics. The inference network had six affine coupling layers. NPE training ran for 100 epochs (1000 iterations per epoch, batch size 32) using the Adam optimizer (initial LR = $5\times 10^{-4}$, cosine decay schedule). The training regime was consistent with typical NPE training regimes for stochastic cognitive models \citep{schumacher2024tutorial}, and the loss value converged by the end of the training.

\begin{table}[t]
\centering
\begin{tabular}{ll}
\toprule
Parameters & Distribution \\
\midrule
$v$ (Drift rate) & $U(0.2,2)$  \\
$a$ (Boundary separation) & $U(0.5,2.5)$  \\
$T_{er}$ (Non-decision time) & $\mbox{Gamma}(1.5,0.2)$  \\
\bottomrule
\end{tabular}
\caption{Prior distributions for the parameter recovery study. $\mbox{Gamma}(1.5,0.2)$ denotes a gamma distribution with shape parameter 1.5 and scale parameter 0.2.}\label{priorpr}
\captionsetup{} 
\end{table}

After training, 500 new data sets were simulated to assess parameter recovery. For Bayesflow, simulated data were passed through the trained summary and inference networks to generate posterior samples. JAGS was run with four chains, each using 1000 adaptation steps, 1000 burn-in steps, and 1000 sampling iterations. To make results comparable, the prior in JAGS was the same as the prior in BayesFlow (Table~\ref{priorpr}). Convergence was assessed via $\hat{R}$ statistics \citep{gelman_inference_1992}, with $\hat{R} < 1.01$ indicating acceptable convergence. For the EZ diffusion model, parameters were derived using the three sufficient summary statistics. Out of 500 data sets, 471 converged in JAGS; the remaining 29 data set was excluded from all methods to maintain comparability. No convergence issue occurred in BayesFlow or EZ diffusion. The parameter recovery plots for all three methods can be found in Figure~\ref{fig:prDDM}, and the simulation-based calibration shows that the estimation from Bayesflow was unbiased (see Figure~\ref{fig:ecdf}).

Performance metrics are reported in Table~\ref{rmsesd}. The root mean squared error (RMSE) and the average marginal posterior standard deviation (SD) across 500 data sets measured the accuracy and precision of three estimation methods. The RMSE is defined as: $\text{RMSE}_{\theta} = \sqrt{\frac{1}{n}\sum(\hat{\theta}_{i}-\theta_{i})^2}$, and its variability was measured by the SD of the mean absolute error $|\hat{\theta_i} - \theta|$ across 500 data sets. We also reported the spread of individual marginal posterior SD values as a measure of posterior contraction.

As expected, BayesFlow and JAGS showed highly similar accuracy and precision across all parameters, with negligible differences in RMSE, average posterior SD, and variability. EZ diffusion was slightly less accurate than these two methods.

\begin{table}[h!]
\centering
\begin{tabular}{c c c c c c c c}
\hline
 & \multicolumn{3}{c}{$\text{RMSE}$} & \multicolumn{1}{c}{ } & \multicolumn{3}{c}{Posterior $SD$} \\
\cline{2-4} \cline{6-8}
 & EZ & BayesFlow & JAGS &  & EZ & BayesFlow & JAGS \\
\hline
$v$ & 0.136 (0.088) & 0.113 (0.073) & 0.115 (0.074) &  &  N/A & 0.112 (0.030) & 0.111 (0.030) \\

$a$ & 0.128 (0.087) & 0.076 (0.051) & 0.081 (0.051) &  &  N/A & 0.076 (0.031) & 0.068 (0.028)\\

$T_{er}$ & 0.045 (0.031) & 0.014 (0.010)  & 0.014 (0.010)&  &  N/A & 0.015 (0.008) & 0.012  (0.006)\\
\hline
\end{tabular}
\caption{Performance comparison between methods based on $\text{RMSE}$ and posterior $SD$. The $RMSE$ and the posterior $SD$ were calculated from estimates of 500 data sets using three distinct methods. For both metrics, the value reported in parentheses represents the standard deviation of that metric's values across the 500 data sets. Since EZ diffusion only provides point estimates, no standard deviation is available for this method.}\label{rmsesd}
\end{table}

\subsection*{Interpreting the Learned Summary Statistics}

Given the agreement between estimation in EZ-diffusion and BayesFlow, it is reasonable to assume that the three learned summary statistics $\boldsymbol{s}_{B}$ should contain the same information as the sufficient summary statistics in EZ diffusion, denoted as $\boldsymbol{s}_{EZ}=(M_{RT}, V_{RT}, P_c)$ (see Figure~\ref{fig:ssvs} in Appendix for a direct mapping). 

To explore their relationship, we retrieved $\boldsymbol{s}_{B}$ and $\boldsymbol{s}_{EZ}$ for the 500 data sets used above, and built a multivariate random forest \citep[implemented in the R package \texttt{MultivariateRandomForest}]{segal_multivariate_2011}. It is a non-parametric machine learning algorithm that captures complex non-linear relationships and makes predictions \citep{segal_multivariate_2011}. We trained a multivariate random forest with 250 sets of $(\boldsymbol{s}_{B}, \boldsymbol{s}_{EZ})$ and used the remaining 250 $\boldsymbol{s}_{B}$ as a test set to predict the corresponding $\boldsymbol{s}_{EZ}$. 

\begin{figure}[!ht]
\centering
\includegraphics[width=1\linewidth]{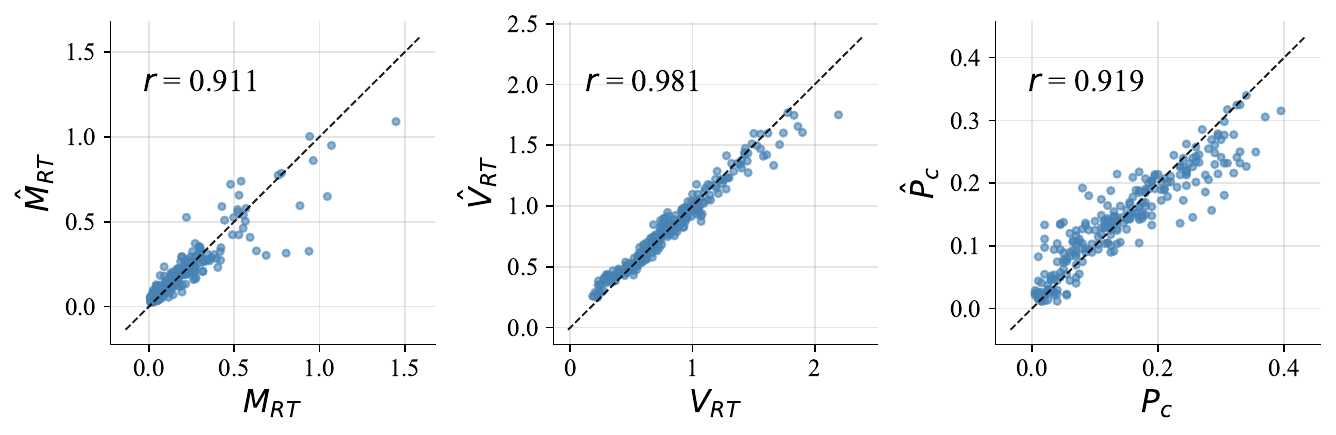}
\caption{\label{fig:ss}Predicted sufficient summary statistics $\hat{\boldsymbol{s}}_{EZ}$ from $\boldsymbol{s}_{B}$. The x-axis are the observed values of $M_{RT}$, $V_{RT}$, and $P_c$, and the y-axis is the $\hat{M}_{RT}$, $\hat{V}_{RT}$ and $\hat{P}_c$ predicted by $\boldsymbol{s}_{B}$}.
\end{figure}

As shown in Figure~\ref{fig:ss}, the correlation coefficient between the true $\boldsymbol{s}_{EZ}$ and $\hat{\boldsymbol{s}}_{EZ}$ predicted by $\boldsymbol{s}_{BF}$ was higher than 0.9. This empirical result is not surprising \citep[]{radev_bayesflow_2020}. Yet, it convincingly demonstrates that summary networks learn a representation of the data that aligns with the information used by analytical methods. 

\section*{The Impact of Outliers in Amortized Bayesian Inference}

Next, we assessed the robustness of ABI with the stylized sensitivity curve \citep[SSC,][]{Andrews1972} and the breakdown point \citep[BP;][]{donoho_notion_1982}. To the best of our knowledge, this is the first study to apply these in the context of ABI.
\subsection*{The Stylized Sensitivity Curve}
When fitting parametric models, the data are assumed to follow a distribution $F_{\btheta}$ with unknown parameters $\btheta$. Bayesian methods return an approximation to the posterior $p(\btheta \mid \bx)$, which can be summarized into a point estimate  $\hat{\btheta}$ (e.g., the posterior mean), and the dependence of the estimator on $\bx$ is denoted as $\hat{\btheta}(\bx)$. If the sample size $n$ becomes very large (i.e., $n \to \infty$), the information in the sample $\bx$ converges to the information provided by the distribution $F$, and thus $\hat{\btheta}(F) \rightarrow \btheta$, assuming the estimator is consistent. However, contamination may be present.
Denoting the contamination distribution as $G$ and the (small) fraction of contaminants as $\epsilon \in [0, 1]$, we can define the \textit{contaminated distribution} $F_{\btheta}^{\epsilon,G}$ as follows:
\begin{equation}
    F_{\btheta}^{\epsilon,G} = (1-\epsilon)F + \epsilon\,G. \label{eq:def outlier} 
\end{equation}
In the study of the influence of outliers, the contaminant distribution is usually defined as a point mass $G \equiv \delta_{x^c}$, with
\begin{equation}
    \delta_{x^c} = \begin{cases}
        1 & \text{if } x=x^c \\
        0 & \text{otherwise,}
    \end{cases} 
\end{equation}
where $x^c$ is the contaminant. Thus, we can define the influence function \citep[IF;][]{maronna_robust_2006} for an estimator $F(\hat{\btheta})$ and a point mass contaminant distribution at $x^c$ as:
\begin{equation}
    \text{IF}_{\hat{\btheta}}(x^c,F) =  \lim_{{\epsilon \to 0}} \frac{\hat{\btheta}\big((1-\epsilon)F + \epsilon \delta_{x^c}\big)-\hat{\btheta}(F)}{\epsilon}.
    \label{eq:def inf fun} 
\end{equation}
The IF indicates how much influence a contaminant $x^c$ exerts on the estimate $\hat{\btheta}$ asymptotically. The IF can be traced for a range of $x^c$ values, resulting in an \textit{influence curve} where the x-axis depicts the value of $x^c$, and the y-axis depicts the influence of $x^c$ on $\hat{\btheta}$.

Finding a closed-form expression for the IF of an estimator of an arbitrary $F$ is hard in practice. Instead, we can compute the stylized sensitivity curve \citep[SSC,][]{Andrews1972} to approximate IF through simulations. The bias a contaminant $x^c$ introduces to an estimate is given by $\hat{\btheta}(\bx, x^c) - \hat{\btheta}(\bx)$, where $\hat{\btheta}(\bx, x^c)$ is the estimate calculated with the original sample $\bx$ plus the contaminant $x^c$ (and thus having n+1 as sample size \footnote{While dividing this bias with $\frac{1}{(n+1)}$ in large samples yields the asymptotic bias of $x^c$, we omit this denominator to allow for sample size dependence.}). Since the bias inherently depends on a specific finite sample $\bx$, using a single simulated sample would introduce random variability. To mitigate this and obtain a more reliable representation, \textit{pseudo data} are being used. Pseudo data consist of the quantiles of the underlying distribution $F$, defined as $x_i = F^{-1}\left( \frac{(i-0.5)}{n} \right)$. When an analytical form for $F$ is unavailable, these quantiles are found by simulating a large number of observations from the observation model to accurately approximate $F$, and then computing the empirical quantiles. This approach ensures a highly representative base sample, free from single-sample randomness, and is called ``stylized''\citep{Andrews1972}.

Furthermore, as the underlying distribution $F$ depends on the true parameter $\btheta$, the precise shape of the SSC0 depends on these parameters. To obtain a reliable estimate of SSC across plausible parameter spaces, one can average the SSCs over $B$ distinct sets of pseudo data, each generated based on different $F_{\btheta}$, with $\btheta$ drawn from a prior or varying across a plausible range:
\begin{equation}
    \overline{SSC}(x^c) = \frac{1}{B} \sum_{b=1}^B \left[ \hat{\btheta}(\bx^b, x^c) - \hat{\btheta}(\bx^b) \right],\label{ei}
\end{equation} with $\bx^b$ being the $b$th pseudo data. By varying the value of $x^c$, we can plot an expected SSC that illustrates the relationship between the $x^c$ value and the impact it has on estimates.

\subsection*{The Breakdown Point}
The BP is the minimum amount of contamination that can lead an estimator to produce extremely aberrant values \citep{donoho_notion_1982}. Assume that $\Theta$ is parameter space for the parameter vector $\btheta$ (i.e., $\btheta \in \Theta$). The asymptotic contamination BP, denoted $\epsilon^\star$, for an estimator $\hat{\btheta}$ at the distribution $F$ is defined as the largest $\epsilon^\star \in [0,1]$ such that for $\epsilon < \epsilon^\star$, $\hat{\btheta}\big((1-\epsilon)F + \epsilon G)$ remains bounded away from the boundary of $\Theta$, regardless of the contaminating distribution $G$ \citep[]{maronna_robust_2006}. 
What this means is that the contamination should not drive the estimated parameter $\hat{\btheta}$ to the boundary of $\Theta$ (or to infinity if $\Theta$ is unbounded for some or all of its components). A high BP means that $\epsilon^\star$ is large and thus that $100\epsilon^\star$ of the sample can be contaminated while leaving the estimate still bounded. For example, the BP of the sample mean (as an estimator of the population mean) is 0\%, while the BP of the median is 50\%.

In a practical setting with a finite sample size $n$, we define a fraction $p=\frac{m}{n}$, so that $m= p \cdot n$ observations in the sample $\bx$ are replaced by $x^c$ (without loss of generality, we replace the first $m$ entries, as the models in this paper assume IID observations). The contaminated sample can be denoted as $\bx^{c(p)}$. This leads to an estimate $\hat{\btheta}(\bx^{c(p)})$. Again, the sample average over $B$ samples can be taken: $\overline{\hat{\btheta}}(x^c,p)$. The BP is then the fraction $p^*$ for which all fractions $p<p^*$ lead to an average estimate that is still acceptable: $|\overline{\hat{\btheta}}(x^c,p)-\overline{\hat{\btheta}}| < \Delta$ (where $\Delta$ is some tolerance). Importantly, in this paper, we only studied point contamination (i.e., $G=\delta_{x^c}$), resulting in an approximate version of BP.

\subsection*{The Impact of Outliers in Amortized Estimation of the Mean of a Normal Distribution}

We first studied the SSC and BP in a toy example, the estimation of the mean $\mu$, where $n$ IID samples were drawn from:
\begin{equation}
    x_1, x_2, \dots, x_n \overset{\text{IID}}{\sim} \mathcal{N}(\mu, 1), \label{eq:dnorm}
\end{equation}
so that $\bx = \left\{ x_1,\dots,x_n \right\}$. 
The parameters $\{ \mu_j\}_{j=1}^J$ were sampled from a prior $\mathcal{N}(0,1)$, and for each $\mu_j$, 10 to 100 samples from $\mathcal{N}(\mu_j,1)$ were generated. The neural network architecture consisted of a Deep Set network with two outputs \citep{zaheer2017deep}, and an inference network comprising two affine coupling layers. The training ran for 10 epochs (4000 iterations \rev{per} epoch, batch size 32) using the Adam optimizer (initial LR = $5\times 10^{-4}$, cosine decay schedule). After training, parameter recovery was performed to confirm that the estimated posterior mean aligned almost perfectly with the analytical posterior mean \citep{gelman2013bayesian}.

\subsubsection*{Stylized Sensitivity Curve of the Amortized Estimator of the Mean}

We assessed the contaminant's impact using the SSC for sample sizes $n \in \{10, 20, 100\}$, with $x^c$ systematically from –100 to 100 in steps of 1. For each $x^c$, 100 parameters and 100 data sets (1000 observations per data set) were simulated. From each of these larger data sets, we derived smaller pseudo data sets $\bx$ by extracting $n$ quantiles using the aforementioned $\frac{i-0.5}{n}$ rule (with $i=1,\dots,n$).  We then introduced the contaminant $x^c$ to these pseudo data sets and computed the average change in estimated $\mu$.

This toy example allows an analytical expression for SSC. Since the prior and likelihood are both normally distributed, the difference in posterior mean between $\bx^c=(\bx,x^c)$ and $\bx$ is:
\begin{align*}
    \hat{\mu}(\bx, x^c) - \hat{\mu}(\bx)  &= \frac{1}{n+2}(x_1 + ... + x_n+x^c) - \frac{1}{n+1}(x_1 + ... + x_n) \\
    &=  \frac{1}{n+2}x^c + \frac{1}{n+2} \sum_{i=1}^n x_i - \frac{1}{n+1} \sum_{i=1}^n x_i. \label{eq:EIF in theory}
\end{align*}

This demonstrates a linear relationship between $x^c$ and the theoretical SSC, with slope $\frac{1}{n+2}$. Since the prior is centered around zero, the intercept also takes the value of zero. The theoretical SSC and that obtained from ABI are shown in the left panel of Figure~\ref{fig:EIFBPnormal}. This indicates that the neural estimator behaves identically to the analytical posterior mean in the presence of an outlier.

\subsubsection*{Breakdown Point of Amortized Mean Estimator}

We estimate the BP using an extreme point contaminant $x^c = -100$, replacing a fraction $p \cdot n$ of observations to create contaminated samples $\bx^{c(p)}$. For each $p$ and $n \in {10, 20, 100}$, we simulate 500 data sets and compute the average estimate $\overline{\hat{\mu}}(\bx^{c(p)})$, which is then plotted against $p$. Again, the BP is available analytically for this simple example:
\begin{align*}
    \hat{\mu}(\bx^{c(p)}) &= \frac{1}{n+1}(p \cdot n \cdot x^c +  \sum_{i=n-p \cdot n}^n x_i) \\
    &= \frac{n}{n+1} p x^c + \frac{1}{n+1} \sum_{i=n-p \cdot n}^n x_i, 
\end{align*}
such that averaging over the data $\bx$ and the parameter $\mu$ yields:
\begin{align*}
    \overline{\hat{\mu}}(\bx^{c},p) &=  \frac{n}{n+1} p x^c 
\end{align*}
Thus, even for the smallest fraction $p=\frac{1}{n}$, the estimate can be made arbitrarily large by choosing an appropriate value for $x^c$.

\begin{figure}[!ht]
\centering
\includegraphics[width=1\linewidth]{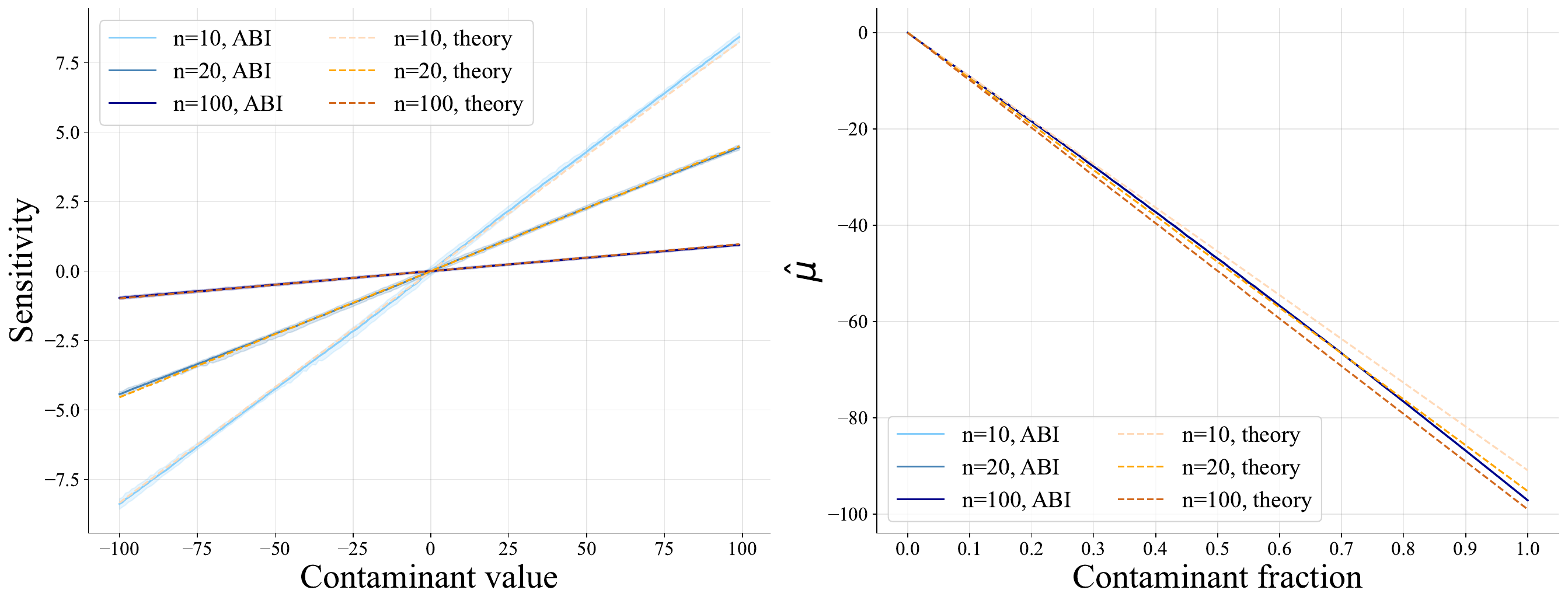}
\caption{\label{fig:EIFBPnormal}The SSC and BP for the $\mu$ estimator when $n \in \{10, 20, 100\}$. The left panel displays the SSC for estimating $\mu$. For reasons of comparison, both the theoretical and the ABI-based SSC are shown. The shaded areas around the ABI-based SSC lines indicate the 2.5\% to 97.5\% percentiles of the influence. The right panel shows the average estimate of $\mu$ with a fraction of $p$ outliers (i.e., copies of $x^c=-100$) inserted into the sample, that is, $\overline{\hat{\mu}}(\bx^{c},p)$.}
\end{figure}

As can be seen from the right panel in Figure~\ref{fig:EIFBPnormal}, the ABI estimation of $\mu$ depended linearly on the fraction of outliers in the sample (as also our derivation has shown). Therefore, it is no surprise that the results from the theory and ABI align closely.

\subsection*{The Impact of Outliers in Amortized DDM Parameter Estimation}

\subsubsection*{Specification of the ABI for Estimating DDM Parameters}

We simulated choice reaction time data to train the NPE, using the priors listed in Table~\ref{priorddm}. The observation model comprised two conditions with opposite drift rates ($v_1$, $v_2$), while other parameters remained fixed in every trial. The number of trials $n$ in each data set varied from 100 to 1000, allowing us, after training, to make inferences on data sets with a wide range of trial numbers. 

The summary network was a Set Transformer \citep[]{lee_set_2019} that output 12 summary statistics, while the inference network is a neural spline flow composed of six coupling layers \citep{durkan2019neural}. The training process consisted of 100 epochs (1000 iterations/ epoch, batch size of 32) using the Adam optimizer (initial LR = $5 \times 10^{-4}$, cosine decay schedule). The loss value converged successfully by the end of the training. Next, we simulated an additional 500 data sets and performed parameter recovery (see row (a) in Figure~\ref{fig:prddmrobust}).

\subsubsection*{Stylized Sensitivity Curve of Amortized DDM Estimator}

We introduced contaminants in the choice reaction time data as a pair ($x^c, y^c$). For each contaminant, the response $y^c \sim \text{Bernoulli}(0.5)$. This randomization served two purposes: (a) it isolated the effect of $x^c$ by averaging out choice influence, and (b) it mirrored the real-world scenario where outliers often arise from random guesses. 

To evaluate our estimator, we systematically varied $x^c$ from 0.01 to 25 seconds in 0.05-second increments. For each $x^c$ value, we generated 100 representative pseudo data sets, each containing 200 trials (100 trials/condition). Our simulation process involved two main steps: (a) parameter and data set simulation: We first simulated 100 parameter sets $(v_1, v_2, a, z, T_{er})$, then simulated 100 full data sets, each consisting of 5000 trials (2500 trials/ condition). (b) Pseudo-data set generation from quantiles: For each simulated full data set, we calculated the error response rates ($e_1$ and $e_2$) for both conditions. These rates allowed us to calculate the approximate number of error trials needed for our pseudo-data sets (rounded to the nearest integer, i.e., $100 \times e_1$ and $100 \times e_2$). We then constructed the 200-trial pseudo-data sets by extracting specific quantiles from the simulated DDM response distributions. For the error trials, we took the first up to the $(100 \times e_1)$th quantile from the error reaction time distribution for condition 1, and similarly for condition 2. For the correct trials, we took the first up to the $(100 - 100 \times e_1)$th quantile from the correct reaction time distribution for condition 1, and similarly for condition 2. Lastly, a contaminating pair $(x^c, y^c)$ was added to each of these pseudo-data sets, specifically within condition 1.

\begin{figure}[!ht]
\centering
\includegraphics[width=1\linewidth]{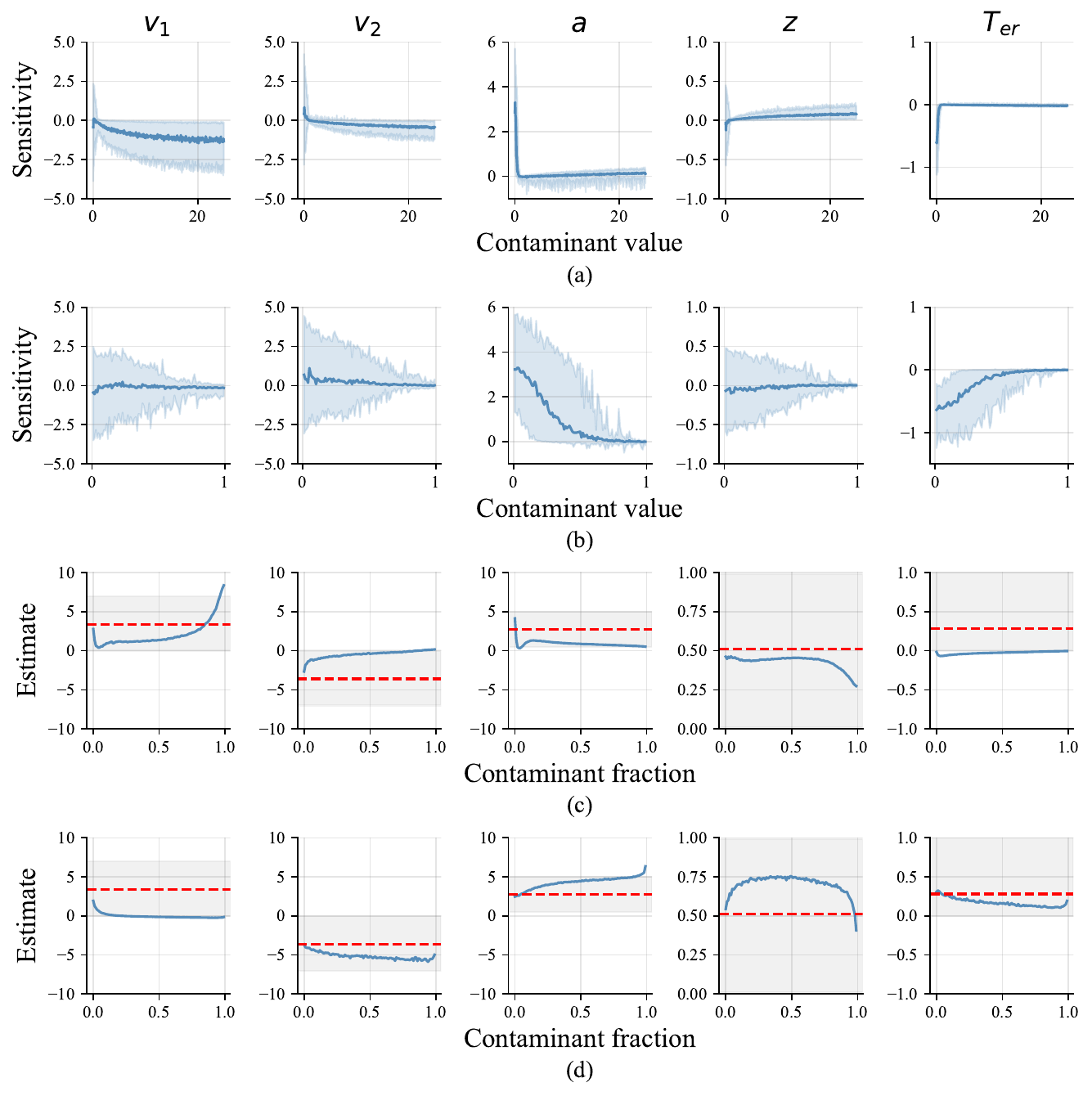}
\caption{SSC and BP of amortized DDM estimation. Row (a) shows the SSC when $x^c$ ranges from 0.01s to 25s. Row (b) zooms in on the SSC when $x^c = 0.01 \sim 1s$, with a higher resolution. Row (c) shows the BP of the estimator when $x^c=0.01s$, while row (d) shows that with large outliers, namely $x^c=20s$. The true posterior mean across 1000 data sets is the prior mean, which is indicated with the red dashed line. The grey area marks the prior density for this parameter. The deviation between the estimates of perturbed data sets (in the blue line) and the red dashed line represents the systematic bias introduced by a certain fraction of outliers.}
\label{fig:EIFDDM}
\end{figure}

Row (a) in Figure~\ref{fig:EIFDDM} shows the SSC for the DDM parameters. To provide a more fine-grained analysis in the short outlier value range, as they are expected to be particularly detrimental, we also plot the SSC of $x^c$ ranging from 0 to 1 second, with an increment of 0.01 seconds as in row (b) of Figure~\ref{fig:EIFDDM}. While the solid blue lines represent the SSC averaged over 100 pseudo-data sets, the shaded areas indicate the 2.5\% to 97.5\% percentiles of the influence. The results showed that even a single contaminant can substantially affect DDM parameter estimates.

When examining the SSC, short contaminants ($<1s$) consistently led to two primary effects: first, an underestimated non-decision time $T_{er}$, and second, an overestimated boundary separation $a$ (see row (b) in Figure~\ref{fig:EIFDDM}). This is likely because an underestimated $T_{er}$ prolongs the ``decision time'', which means more information is needed to reach a decision. While these patterns remained consistent across all 100 pseudo-data sets, the contaminant's impact on drift rates and starting point varied, possibly due to the random sampling of $y^c$. Longer contaminants ($>1s$) have less of an effect on $T_{er}$. While they still affect other parameters—particularly the drift rate $v$ of the first condition—since the altered trial extends the mean reaction time of the first condition, again implying slower evidence accumulation. Overall, both short and long outliers significantly distort DDM estimates.

\subsubsection*{Breakdown Point of Amortized DDM Estimator}

To explore the BP of the NPE, we introduced a fraction $p$ of extreme contaminants $x^c$ into the first condition, assigning random responses $y^c \sim \text{Bernoulli}(0.5)$. For each $p$, 500 data sets (100 observations per condition per data set) were simulated and fitted. We assessed the estimates by comparing the average posterior mean from perturbed data sets to that from the original data set. When the estimate converges to the prior mean (Table~\ref{priorddm}), shown as the red dashed line in rows (c) and (d) in Figure~\ref{fig:EIFDDM}, the estimator is considered to produce reasonable estimates; otherwise, it is considered to have broken down.

Rows (c) and (d) in Figure~\ref{fig:EIFDDM} show the BP results for short outliers ($x^c=0.01s$) and long outliers ($x^c=20s$), respectively. With more $x^c = 0.01s$ in the first condition, the drift rates in both conditions and the boundary separation approached zero. With more $x^c = 20s$ in the first condition, $v_1$ was estimated to be near zero, $v_2$ becomes more extreme, and the boundary separation approaches the upper boundary of its prior. In both cases, the estimation became severely distorted starting at $p^*=\frac{1}{n}$, which was identified as the BP. Notably, multiple short outliers even yielded negative posterior means for $T_{er}$, exceeding its valid range (row (c) in Figure~\ref{fig:EIFDDM}). These results highlight the estimator’s high sensitivity to contamination.

\section*{Robust Amortized Bayesian Inference: Training with Perturbed Data}

Our results show that estimators trained on data strictly conforming to model assumptions were highly sensitive to contamination. However, robustness can be improved by including contaminants in the training data. Throughout the remainder of this paper, we refer to the estimator trained on clean data as the $\textit{standard estimator}$ and the one trained with contaminated data as the $\textit{robust estimator}$. While this approach is conceptually simple, key questions remain, such as (a) whether it truly enhances robustness and (b) how different contamination types affect parameter estimates. We address these questions using both a toy example and the DDM.

\subsection*{Robust Amortized Estimation of the Mean of a Normal Distribution}

Training data of robust estimators $\bx^c$ in the toy example came from a contaminated observation model: 
\begin{equation}
    \bx^c \sim\begin{cases}
		\mathcal{N}(\mu,1), & \text{with probability } 1-\pi\\
        t_{\nu}(\mu,1), & \text{with probability } \pi,
		 \end{cases} \label{eq:robustdnorm} 
\end{equation}
where $t_{\nu}(\mu,1)$ refers to a $t$ distribution with location $\mu$, scale 1, and degrees of freedom $\nu$, and $\pi$ is a predefined contamination probability that we have set to 0.1. In the remainder, we train three estimators with $\nu \in \{1, 3, 5\}$, respectively.

Three robust estimators were trained using the same settings as the standard estimator. After training, 500 new data sets were simulated to evaluate parameter recovery for each estimator.

\begin{figure}[t]
\centering
\includegraphics[width=1\linewidth]{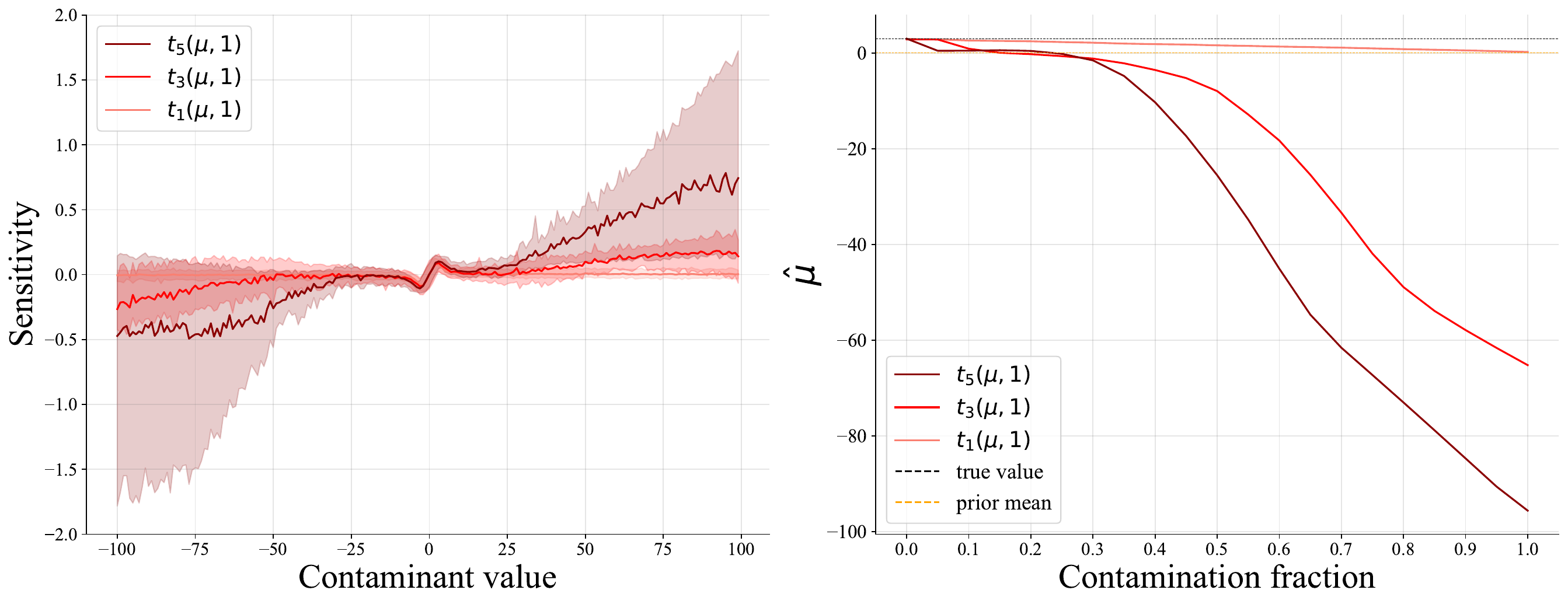}
\caption{SSC and BP of robust neural estimators of $\mu$ trained with different degrees of freedom $\nu$.}
\label{fig:EIFBPnormalrobust}
\end{figure}

Robustness of the estimators was evaluated using SSC and BP, following the same procedure as for the standard estimator ($n=20$). Results are shown in the left panel of Figure~\ref{fig:EIFBPnormalrobust}, to be compared with Figure~\ref{fig:EIFBPnormal} (note the differing y-axis scales). The shaded areas around SSCs indicate the 2.5\% to 97.5\% percentiles of the influence in the 500 data sets. Near the origin, the SSC shows a linear dependence on $x^c$; for more extreme values, the influence is down-weighted. The $t_1$ robust estimator effectively suppressed influence across the entire range, while $t_3$ and $t_5$ estimators show increased sensitivity to extreme outliers—likely due to extrapolation, as such values were absent during training. They also exhibited greater variability in the sensitivity across 100 pseudo data sets, as evidenced by the significantly wider shaded areas for the $t_3$ and $t_5$ compared to the $t_1$ estimator. Table~\ref{outlier} lists the proportion of regular and far outliers (per definition in Tukey (\citeyear{tukey1977exploratory})) for each contamination distribution. Under 10\% contamination, the $t_1$ distribution produces substantially more regular and far outliers than the others.

\begin{figure}[t]
\centering
\includegraphics[width=1\linewidth]{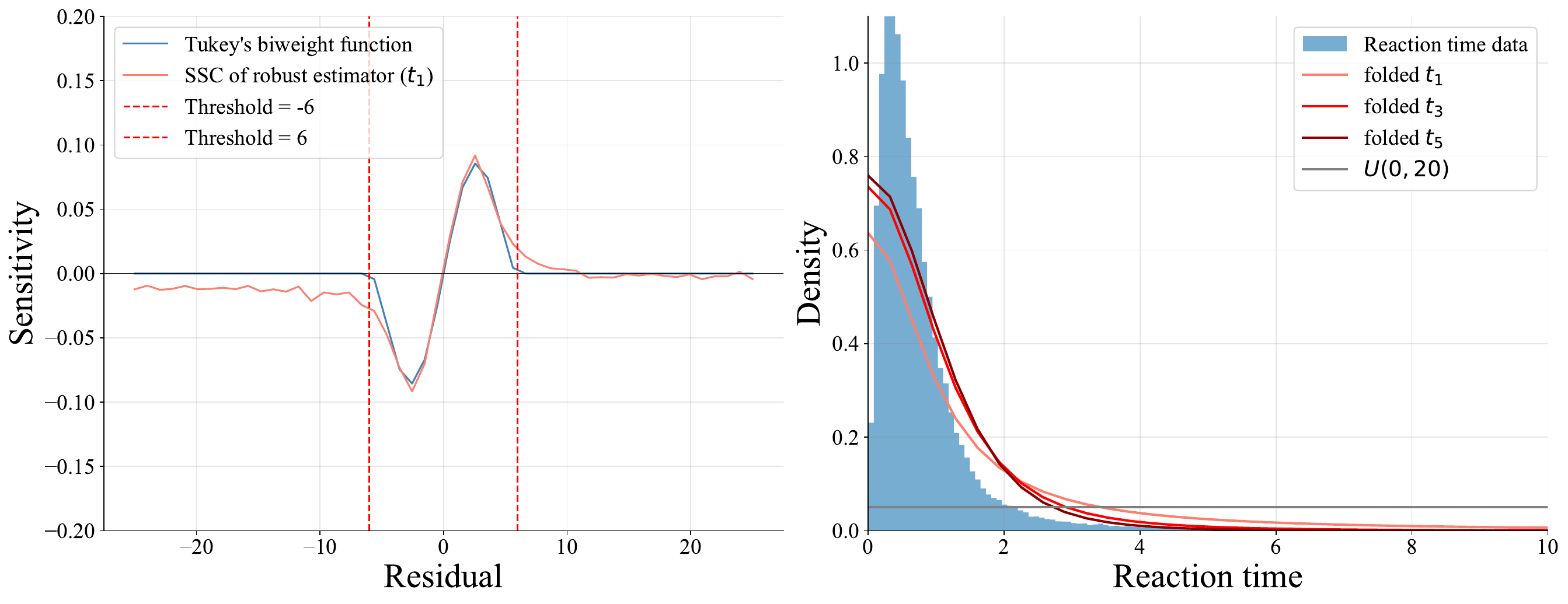}
\caption{Tukey's biweight function and reaction time data. The left panel overlays the SSC of Tukey's biweight loss function and that of the robust $\mu$ estimator with $t_1$. The right panel shows the histogram of reaction times generated by the DDM for 500 prior predictive realizations, overlaid with the densities of the four contamination distributions.}
\label{fig:tukeyandrt}
\end{figure}

Interestingly, the behavior of the neural robust $t_1$ estimator aligned with Tukey's biweight loss function from robust statistics \citep[]{tukey_robust_1979}. The IF of Tukey's biweight loss function is:
\begin{equation}
    \text{IF}_{\hat{\mu}}(x^c,F)= \begin{cases}
		(1-(x^c/k)^2)^2, & \text{if } |x^c| \leq k\\
        0 & \text{if } |x^c| > k,
		 \end{cases} \label{eq:tukeybiweight}
\end{equation}
where $k$ is a tuning constant or a cutoff value that ensures the estimator is insensitive to contaminants in absolute value larger than $k$. When we overlaied the graph of the robust $t_1$ estimator's SSC with Tukey's biweight loss function (using $k=6$), as seen in the left panel of Figure~\ref{fig:tukeyandrt}, their significant overlap suggested the NPE functioned similarly to traditional robust estimators developed under the MLE framework.

We also assessed the BP of the robust estimators by contaminating data sets with increasing fractions $p$ of outliers ($x^c = -100$) and computing the average estimate $\hat{\mu}$ across 500 data sets per estimator. To obtain meaningful BP comparisons, we set the true mean of the validation data sets to be $\mu = 3$ rather than using a prior centered at zero. Otherwise, reversion to the prior mean during breakdown would go undetected. An estimator that is robust to contaminants should show an average estimated mean $\hat{\mu}$ close to the true mean $\mu=3$. As shown in the right panel of Figure~\ref{fig:EIFBPnormalrobust}, the $t_1$-trained estimator demonstrates the greatest resistance, with robustness decreasing as $\nu$ increases. When all values were replaced with $x^c = -100$, the estimates converged to zero (i.e., the prior mean), indicating the network no longer extracted information from the data.

\begin{table}[!ht]
\centering
\begin{tabular}{llcccc}
\hline
          &       & \multicolumn{2}{c}{Regular outliers} & \multicolumn{2}{c}{Far outliers} \\
Estimator &       & $Q_3+1.5\cdot IQR$      & \%         & $Q_3+3\cdot IQR$     & \%        \\ \hline
Robust    & normal + 10\% $t_1$ & 2.828                   & 2.585      & 4.949                & 1.269     \\
          & normal + 10\% $t_3$ & 2.734                   & 1.280      & 4.785                & 0.174     \\
          & normal + 10\% $t_5$ & 2.719                   & 1.008      & 4.758                & 0.051     \\
Standard  &  normal     & 2.698                   & 0.698      & 4.721                & 0.0002    \\ \hline
\end{tabular}
\caption{Percentage of regular and far outliers under three contaminated simulators (with $\pi=.1$) and the standard normal simulator. Regular outliers are data points beyond 1.5 times the interquartile range (IQR) from the lower and upper quartiles (i.e., $Q_1$ and $Q_3$, respectively), while far outliers are those beyond 3 times the IQR from the quartiles \citep[]{tukey1977exploratory}. This table shows the percentage of regular and far outliers in different simulators, along with the corresponding thresholds. Only the upper boundaries are shown because the distributions we use in simulators are symmetric.}\label{outlier}
\end{table}

Overall, assuming a contaminated observation model when training the NPE significantly enhanced the robustness in estimating $\mu$, and $t_1$ performed best among our candidate contamination distributions.

\subsection*{Robust Amortized DDM Estimation}

For robust DDM estimation, we produced contaminated data $(\bx^c, \by^c)$ as follows:
\begin{equation}
    (\bx^c, \by^c) \sim \begin{cases}
     \text{Wiener}(v,a,z,T_{er}), & \text{with probability } 1-\pi \\
     G, & \text{with probability } \pi
\end{cases} 
\end{equation}
with $\pi = 0.1$. We tested four different contamination distributions $G$: $x^c$ was drawn from a \text{folded-}$t_{\nu}$ distribution (i.e., the absolute value of $t$ distributed random variable) or from a uniform distribution $U(0,20)$:
\begin{align*}
    x_i^c \sim \begin{cases}
        \text{folded-}t_{1} & \text{ or } \\
        \text{folded-}t_{3} & \text{ or } \\
        \text{folded-}t_{5} & \text{ or } \\
         U(0,20). &
    \end{cases}
\end{align*}
The corresponding response was also contaminated $y_i^c \sim \text{Bern}(0.5)$, implying random guessing.

The right panel of Figure~\ref{fig:tukeyandrt} shows the histogram of perfectly clean reaction time data (for a variety of parameter vectors; in blue) overlaid with the density of the different contamination distributions.

Four robust estimators were trained using the exact sample sizes, network architecture, and training epochs as the standard DDM estimator. All robust estimators converged and accurately recovered parameters (see row (b) to (e) in Figure~\ref{fig:prddmrobust}).

\begin{figure}[!ht]
\centering
\includegraphics[width=0.99\linewidth]{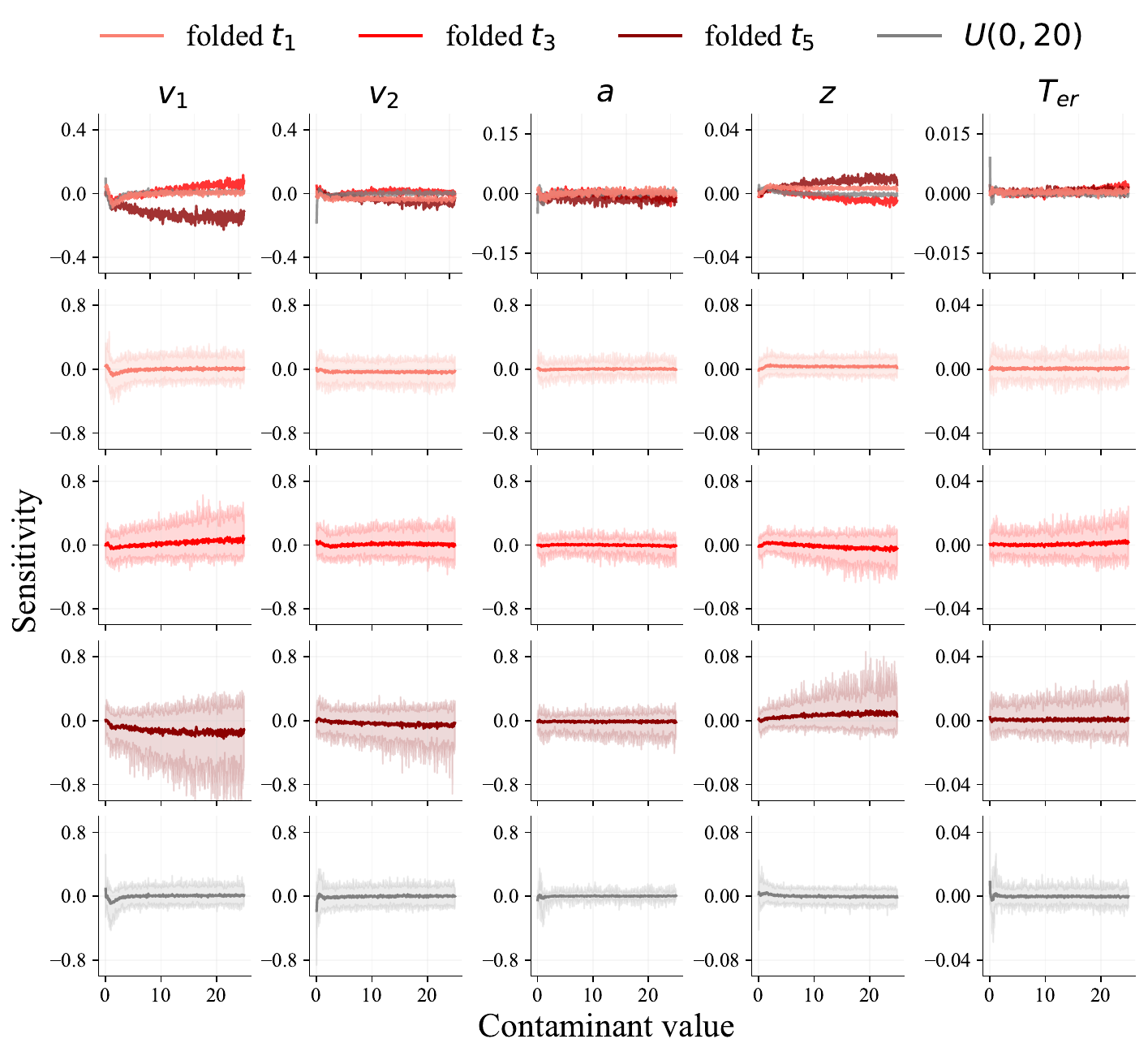}
\caption{\label{fig:sscddmrobust} The SSC of robust DDM estimators with $x^c$ ranges from 0 to 25s. The first row provides a comparative overview of the SSC for four distinct robust estimators. The last four rows show the variability in sensitivity across the 100 pseudo-data sets in four robust estimators, respectively.}
\end{figure}

\begin{figure}[!ht]
\centering
\includegraphics[width=0.99\linewidth]{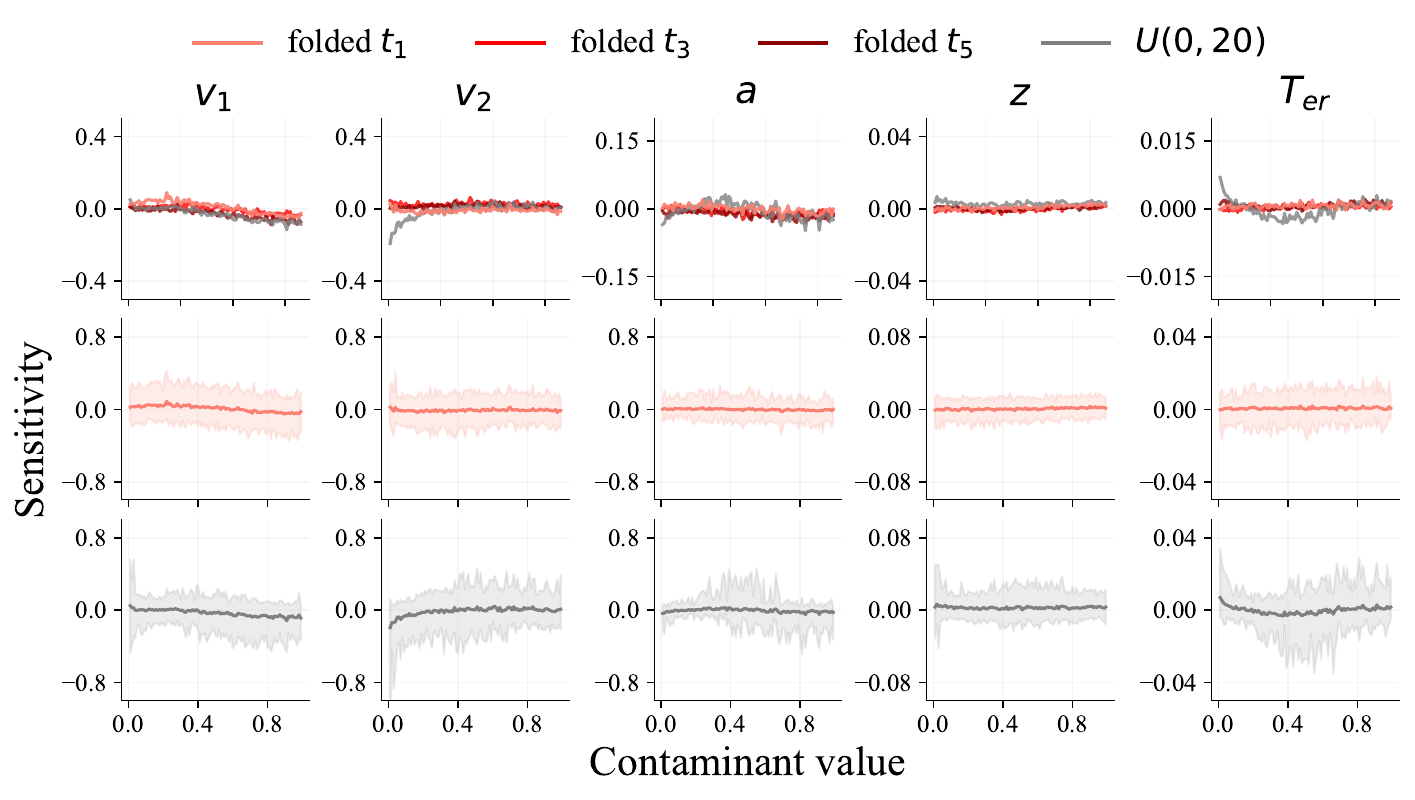}
\caption{\label{fig:sscddmrobust01} The SSC of robust DDM estimators with $x^c$ ranges from 0 to 1s. The first row provides a comparative overview of the SSC for four distinct robust estimators. The second row is the variability in sensitivity across 100 pseudo-data sets in the robust estimator with folded $t_1$, and the third row shows that in the $U(0,20)$ estimator. The plots for the folded $t_3$ and $t_5$ robust estimators are omitted, as they are highly similar to the folded $t_1$.}
\end{figure}

We then evaluated the performance of robust DDM estimators using the SSC. Overall, Figure~\ref{fig:sscddmrobust} shows that all four robust estimators demonstrate resistance to both short and long outliers (Figure~\ref{fig:sscddmrobust} should be compared to Figure~\ref{fig:EIFDDM}). Among all the $t$-based robust estimators, a lower $\nu$ resulted in: (a) average sensitivity closer to zero, indicating higher robustness, (b) lower variability in sensitivity across 100 pseudo-data sets. This means that a lower $\nu$ led to reduced sensitivity to contaminants, especially when $x^c$ was large. Compared to the robust estimator with folded $t_1$, the estimator with $U(0, 20)$ showed greater sensitivity towards short outliers between $0-1$s, and the variability in estimated difference was higher (Figure~\ref{fig:sscddmrobust01}). This is possibly because the $t$-based robust estimator assigned higher weights to short contaminants than the $U(0, 20)$ robust estimator (see the right panel in Figure~\ref{fig:tukeyandrt}). Yet, interestingly, the $U(0, 20)$ robust estimator exhibited resistance to out-of-training $x^c$ values (e.g., 20s to 25s), suggesting it effectively learned to filter out large values.

\begin{figure}[!ht]
\centering
\includegraphics[width=0.99\linewidth]{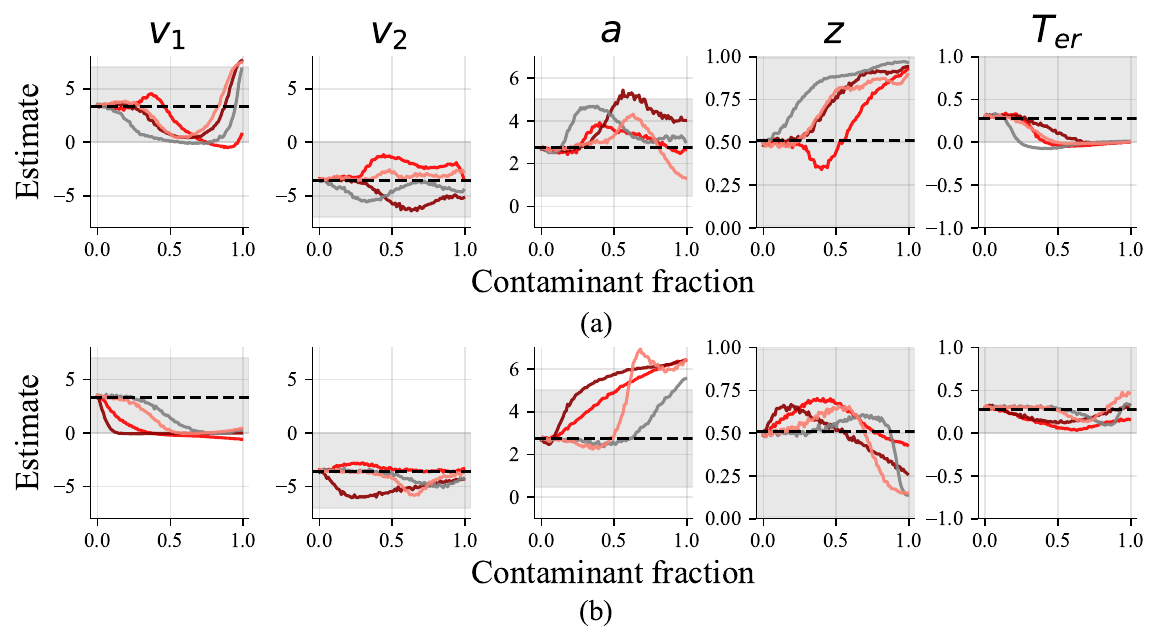}
\caption{BP of robust DDM estimators (applied to data sets with two conditions and 100 trials per condition). Panel (a) displays the BP of the estimator when $x^c=0.01s$, while panel (b) shows the BP with large outliers ($x^c=20s$). The true posterior mean across 1000 data sets is indicated with the black dashed line. The grey area marks the prior density for this parameter. The deviation between the estimates of perturbed data sets and the black dashed line is the systematic bias that a certain fraction of outliers brings.}
\label{fig:eifbpddmrobust}
\end{figure}

The breakdown properties of the robust estimators are further illustrated in Figure~\ref{fig:eifbpddmrobust}, using $x^c = 0.01$s and $x^c = 20$s across contamination levels $p \in [1/100, 100/100]$ in the first condition. When $x^c = 0.01s$, the uniform estimator broke down earlier than the $t$-based estimators (see the upper panel in Figure~\ref{fig:eifbpddmrobust}). This difference arose because the folded-$t$ distributions assigned more weight near zero than the $U(0, 20)$. Within the $t$-based robust estimators, there was no significant difference in BP, and the average estimate deviated from the prior mean when there was around 25\% contamination. This was because the weights that folded- $t_1$, $t_3$, and $t_5$ assigned to short contaminants were similar. When $x^c = 20s$, the robust estimator with $U(0, 20)$ exhibited the highest BP. Among all the $t$-based estimators, the lower the $\nu$, the higher the BP, as lower $\nu$ indicates higher weights on long contaminants. Overall, for DDM, the $t_1$-based estimator offered the highest BP (approximately 25\%) in both short and long point contamination scenarios.

In summary, training with a contaminated observation model improved the robustness of DDM estimators. The choice of contamination distribution had a critical impact on robustness. For $t$-based robust estimators, lower $\nu$ led to higher robustness and less variability in the influence from one contaminant, especially for large outliers. Meanwhile, the $t_1$-based estimator generally offered the most robust performance across both short and long point contamination scenarios. The $U(0,20)$-based estimator showed higher sensitivity to short contaminants, while exhibiting resistance to large, out-of-training contaminants. It broke down early for short outliers but displayed the highest BP for long ones. Based on these results, the $t_1$-based robust estimator is considered the best candidate for contamination distribution, and additional exploratory simulations are presented in the Appendix, varying the contamination probability $\pi$ under the $t_1$ distribution.

\section*{The Cost of Robustness}
It is well understood that any robust approach comes with costs \citep{maronna_robust_2006}. For example, robust M-estimators often exhibit efficiency losses, which refer to a higher asymptotic variance of the robust estimator compared to the standard estimator in uncontaminated data estimation \citep{hampel_robust_2005}. We explored the cost of the robust NPE, evaluating (for the $j$th parameter $\theta_j$) the accuracy and efficiency of a robust estimator $\hat{\theta}_j^R$ compared to a standard estimator $\hat{\theta}_j^S$. 

As a measure of accuracy, we computed the ratio of mean absolute error (MAE) as follows: 
\begin{equation}
   \text{MAE Ratio}^{S,R}_j =\frac{1}{B} \sum_{b=1}^B \frac{|\hat{\theta}_j^S(\bx_b)- \theta_{j(b)}|}{|\hat{\theta}_j^R(\bx_b)- \theta_{j(b)}|}, \label{eq:absolutebias} 
\end{equation}
where $\bx_b$ is the $b$th data set simulated from the standard model, $\theta_{j(b)}$ is the true $j$th parameter value for $b$th data set, and $B$ is the total number of data sets.

For efficiency loss, we used the ratio of average posterior variance:
\begin{equation}
   \text{Posterior Variance Ratio}^{S,R}_j =\frac{1}{B} \sum_{b=1}^B \frac{\mbox{var}(\hat{\theta}_j^S \mid \bx_b)}{\mbox{var}(\hat{\theta}_j^R \mid \bx_b)}, \label{eq:psd}
\end{equation}
where $\text{var}(\hat{\theta}_j \mid \bx_b)$ is the variance of the marginal posterior distribution of the $j$th parameter based on the $b$th simulated data set $\bx_b$.

\subsection*{The Cost of Robustness in the Estimation of the Mean}

We simulated 20,000 clean data sets ($n=20$) and fit them using three robust estimators, comparing each to the standard estimator. The MAE ratio between $t_1$, $t_3$, $t_5$ and the standard estimator is 0.999, 0.998, 0.999, respectively, and the posterior variance ratio is 0.909, 0.943, and 0.968, respectively. The MAE ratios were close to 1, indicating minimal accuracy loss (0.1–0.2\%). However, variance ratios revealed notable efficiency loss, with the $t_1$-based estimator showing the largest (9.0\%). This aligned with the efficiency loss observed in M-estimators for finite samples ($n=20$) when estimating $\mu$ in a univariate normal distribution \citep{wu1985robust}. For example, Hampel, Huber, and Tukey's biweight ($k=6$) M-estimators exhibited efficiency losses of 4.8\%, 17.0\%, and 11.2\% compared to the optimal least-squares estimator, respectively. Our robust estimators with $t_1$, $t_3$, and $t_5$ showed 9.0\%, 5.5\%, and 3.3\% efficiency losses compared to a standard neural $\mu$ estimator, respectively. This highlights the $t_1$ estimator's strong robustness with reasonable efficiency. As $\nu$ increases, efficiency improves—illustrating the trade-off between robustness and efficiency, consistent with SSC/BP results in Figure~\ref{fig:EIFBPnormalrobust}.

\subsection*{The Cost of Robustness in DDM Estimation}

We simulated 10000 data sets ($n=100$) and calculated the MAE ratios and posterior variance ratios compared to the standard estimator. The results are shown in Tables~\ref{maeddm} and~\ref{vddm}, respectively. As expected, the accuracies were lower for all the robust estimators (e.g., depending on the parameter, about 12\% to 24\% loss in accuracy for the robust estimator based on the $t_1$ distribution) compared to the standard one. No systematic relationship between $\nu$ and the performance of the estimators was detected. The robust estimator with $U(0,20)$ exhibited less accuracy loss (from around 6\% to 15\% for the five parameters). 

All robust estimators showed moderate to high efficiency loss, with up to 43\% for $T_{er}$ in the folded-$t_1$ model. However, this loss is not directly comparable to that in estimating $\mu$ due to the DDM's greater complexity and parameter interdependence. The robustness–efficiency trade-off can be adjusted by lowering the contamination probability $\pi$ during training, reducing efficiency loss (see Table~\ref{tradeoffnormal},\ref{tradeoffMAE}, and \ref{tradeoffVAR}). However, as shown in Figure~\ref{fig:robustpi} and the Appendix, decreasing $\pi$ also lowered the estimator’s BP.

\begin{table}[!ht]
\caption{MAE ratios of the four robust neural DDM estimators. Each row compares the robust estimator to the standard estimator. \label{maeddm}}
\begin{threeparttable}
\begin{tabular*}{\columnwidth}{@{\extracolsep{\fill}}lccccc@{\extracolsep{\fill}}}
\toprule
Robust estimator & $v_1$ &  $v_2$ & $a$ & $z$ & $T_{er}$ \\
\midrule
$t_1$ & 0.76  & 0.79 & 0.769 & 0.88 & 0.792  \\
$t_3$ & 0.731 & 0.722 & 0.728 & 0.865 & 0.764  \\
$t_5$ & 0.728 & 0.74 & 0.734  & 0.862 & 0.733  \\
$U(0,20)$ & 0.876 & 0.894 & 0.907 & 0.936 & 0.848\\
\bottomrule
\end{tabular*}
\end{threeparttable}
\end{table}

\begin{table}[!ht]
\caption{Posterior variance ratios of the four robust neural DDM estimators. Each row compares the robust estimator to the standard estimator.\label{vddm}}
\begin{threeparttable}
\begin{tabular*}{\columnwidth}{@{\extracolsep\fill}lccccc@{\extracolsep\fill}}
\toprule
Robust estimator & $v_1$ &  $v_2$ & $a$ & $z$ & $T_{er}$  \\
\midrule
$t_1$ & 0.613 & 0.596 & 0.621 & 0.732 & 0.571 \\
$t_3$ & 0.637 & 0.569 & 0.634 & 0.694 & 0.581  \\
$t_5$ & 0.618 & 0.569 & 0.672 & 0.713 & 0.568  \\
$U(0,20)$ & 0.723 & 0.672 & 0.723 & 0.756 & 0.636\\
\bottomrule
\end{tabular*}
\end{threeparttable}
\end{table}
 
\section*{Real Data Example}

To demonstrate the practical utility of our robust approach, we applied it to the data set from \cite{ratcliff_modeling_1998}, available as \verb+rr98+ in the \verb+RTDists+ R package \citep{rtdists2022}. We fit both the raw and outlier-cleaned versions of the data using standard and robust estimators to compare their estimates.

In this task, three participants were asked to decide whether the pixel arrays were ``bright'' or ``dark''. The experiment was a $33 \times 2$ within-subject design: brightness strength (0–100\% white pixels across 33 levels) and instruction (speed vs. accuracy). This setup targeted the speed–accuracy trade-off in decision-making. Each participant completed approximately 4,000 trials per instruction condition. To simplify analysis, the 33 brightness levels are grouped into five bins. we fit separate DDMs for the speed and accuracy instructions as they may differentially affect boundary separation and other parameters. This follows the approach of Ratcliff and Rouder (\citeyear{ratcliff_modeling_1998}) and Singmann (\citeyear{singmann_reanalysis_2022}).

Figure~\ref{fig:ds} shows the proportion of ``dark'' responses across brightness bins, revealing a clear effect of stimulus strength on choice. We also plot RT quantiles (10\%, 30\%, 50\%, 70\%, 90\%) by strength level and instruction condition. As expected, the speed condition yielded shorter RTs, highlighting substantial differences between conditions (Figure~\ref{fig:quantile}).

The standard estimator was trained on clean reaction time data, while the robust estimator used a $\text{folded-}t_1$ contamination model with a 10\% contamination probability. Both used identical network architectures and training parameters as in the previous section. The number of observations per data set in the training data varied from 100 to 5,000 trials to maintain flexibility for data sets up to 4,000 trials. Training was performed on an NVIDIA Tesla V100-SXM2-32GB GPU, with each epoch taking 95 seconds and a total training time of 158 minutes per network.

Both raw and cleaned data sets were input to the two networks. Data cleaning followed the original procedure by Ratcliff and Rouder (\citeyear{ratcliff_modeling_1998}), excluding trials with reaction times below 200ms or above 2500ms.

\begin{figure}[!ht]
\centering
\includegraphics[width=1\linewidth]{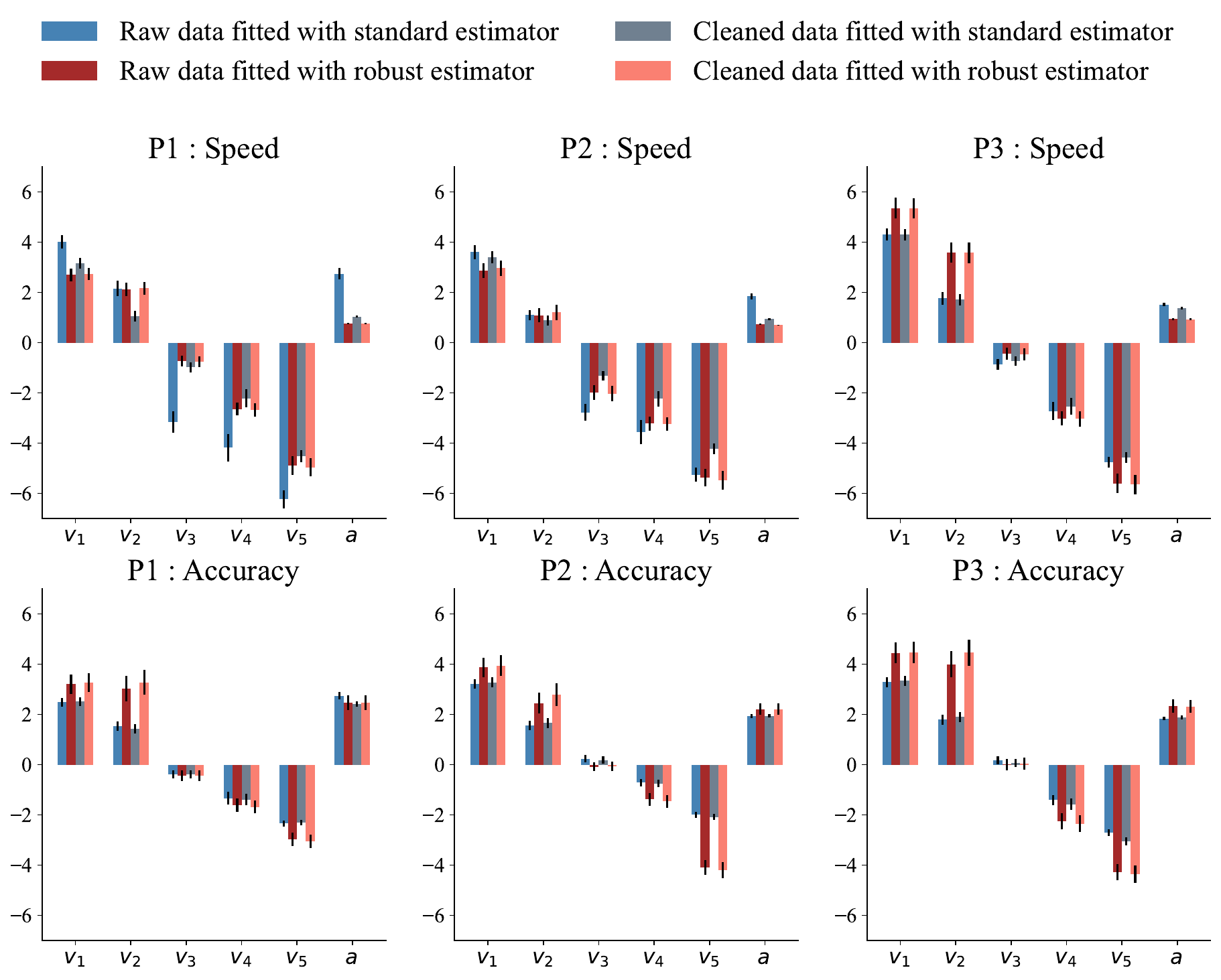}
\caption{Inference of drift rate and boundary separation across different data sets and estimators. The y-axis represents the estimated values, while the x-axis shows the parameter names. The error bar represents one posterior standard deviation. These parameters are displayed in a single figure due to their similar scales.}\label{fig:va}
\end{figure}

\begin{figure}[!ht]
\centering
\includegraphics[width=1\linewidth]{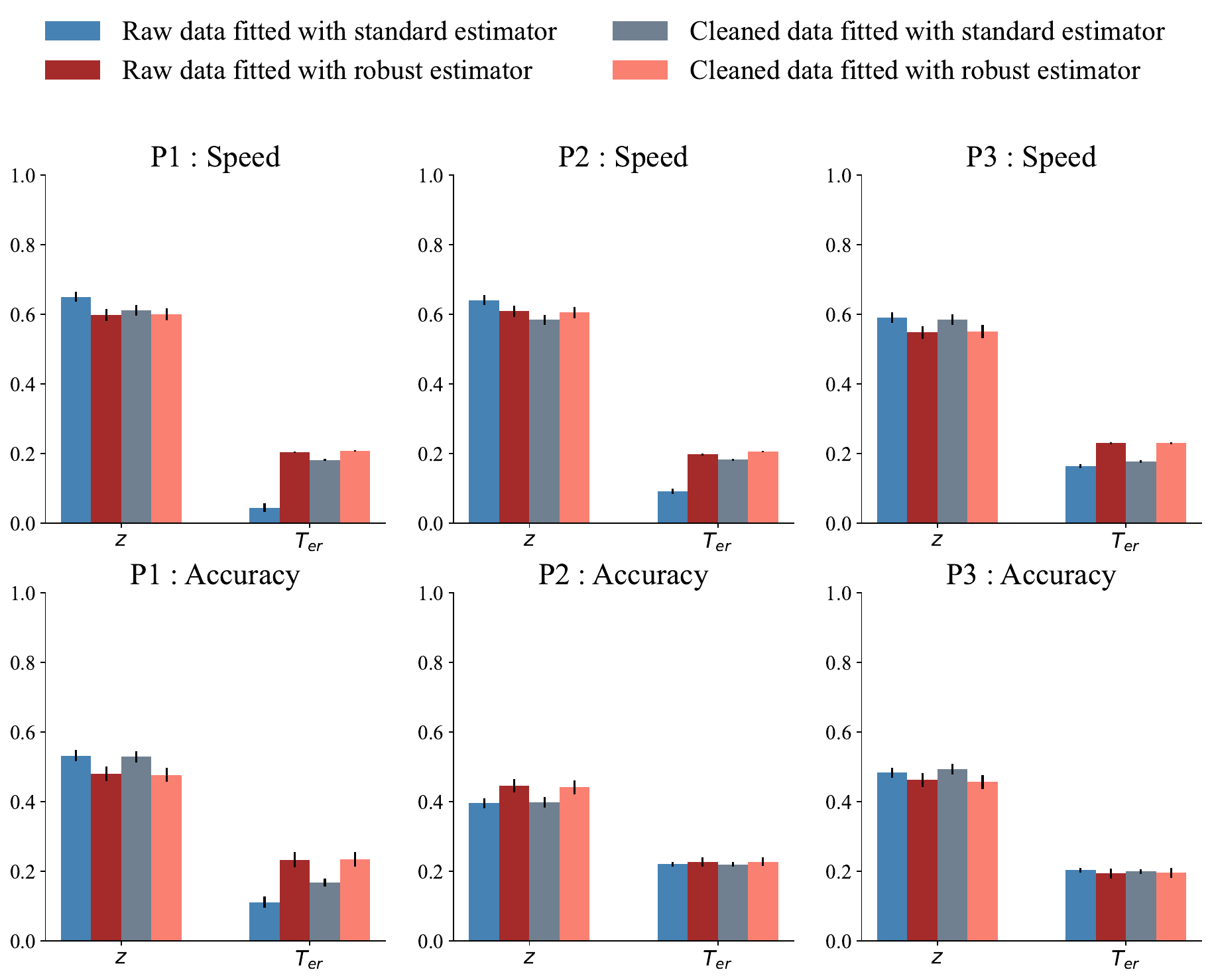}
\caption{Inference of response bias and non-decision time across different data sets and estimators. The y-axis represents the estimated values, while the x-axis shows the parameter names. The error bar represents one standard deviation in the posterior distribution. These parameters are displayed in a single figure due to their similar scales.}\label{fig:zt}
\end{figure}

Posterior means and standard deviations for all conditions are shown in Figures~\ref{fig:va} and~\ref{fig:zt}. Under the speed instruction, fitting raw versus cleaned data with the standard estimator yielded notable differences, particularly for participants 1 and 2. In contrast, for the accuracy condition, the impact of data cleaning was minor. This is likely because most outliers in the accuracy condition are long (95\% between 2.52s and 6.98s), which may reflect genuine decision hesitations rather than contamination. Long outliers generally have less influence on parameter estimates than short ones. Under the speed condition, outliers are predominantly short (95\% between 0.123s and 0.199s), which strongly affects estimation. In contrast, the robust estimator produced consistent results regardless of whether raw or cleaned data were used.

Since real data lack ground-truth parameters, we assessed how well the estimates reflect experimental design effects. Parameters $v_1$ to $v_5$ correspond to conditions with increasing proportions of white pixels, with $v_1$ representing the lowest and $v_5$ the highest. Given that the ``dark'' response is the upper bound, we expected an ordinal trend: $v_1 > v_2 > v_3 > v_4 > v_5$ under both speed and accuracy instructions, with $v_3$ nearest to zero due to maximal perceptual noise. This expected pattern was successfully recovered by the robust estimator applied to raw data.

As for other parameters, in the speed condition, the standard estimator applied to raw data severely underestimated non-decision time for participants 1 and 2 due to short outliers. This underestimation biased other parameters, notably inflating drift rates and boundary separation (see row (b) in Figure~\ref{fig:EIFDDM} and Figure~\ref{fig:va}). The bias in boundary separation was obvious when comparing speed and accuracy conditions. While speed instructions are typically associated with lower boundary separation \citep{ratcliff_theory_1978, ratcliff_modeling_1998, ratcliff_diffusion_2016}, this expected pattern emerged only in the robust estimates—not in the standard estimator, which failed to capture the effect, potentially leading to incorrect conclusions.

Overall, the robust estimator effectively recovered expected patterns induced by experimental manipulations, such as the ordinal ranking of conditions, while mitigating biases from outliers. In contrast, the standard estimator was sensitive to outliers, often underestimating non-decision time and inflating other parameters, which can lead to misleading conclusions. Notably, the robust estimator under the ABI framework demonstrated strong performance and practical accessibility for real-world applications.

\section*{Discussion}
This paper focused on testing and improving the robustness of ABI in the context of cognitive modeling. 
It provided the first systematic exploration of the robustness of ABI, combining theoretical analysis and empirical assessment. It also offered a straightforward method for robustifying ABI, increasing its reliability across a wide range of applications in psychology.

The parameter recovery study showed that ABI can perform posterior inference for stochastic cognitive models to the same level of accuracy and precision as MCMC sampling. Additionally, the comparison between learned summary statistics and sufficient summary statistics being used in EZ diffusion \citep[]{wagenmakers_ez-diffusion_2007} uncovered the black box of neural networks. These findings position ABI as a method that can maintain interpretability while leveraging the flexibility and capacity of neural networks.

The impact of contaminants in ABI was then assessed with SSC and BP. When estimating $\mu$ in a univariate normal distribution, the SSC and BP in analytical posterior inference and ABI nicely coincided. This congruence is promising as it implies that ABI does not introduce unforeseen or erratic reactions to the presence of contaminants. In DDM estimates, our analysis showed that even a single contaminant led to bias, which pointing out the necessity of improving the robustness of ABI to ensure accuracy in contamination-prone data sets.

To enhance the robustness of ABI, a simple yet effective data augmentation method was proposed. By assuming a fraction of the training data (e.g., 10\%) comes from the contamination distribution, the neural networks can learn to map contaminated data to the correct posterior. We examined contamination distributions based on $t$ distributions with different degrees of freedom $\nu$ and a uniform distribution. In both toy example and DDM estimation, we found that assuming contamination arising from a $t$ distribution with one degree of freedom (i.e., a Cauchy) largely improved robustness. Meanwhile, the robust $\mu$ estimator with Cauchy distribution exhibited the same SSC as Tukey's Biweight function \citep{tukey_robust_1979}, an M-estimator that applies transformations on residuals when using the traditional least squares method. These results suggest that the data augmentation approach can achieve similar effects to manually manipulating the loss function. This method is a very straightforward approach that addresses a crucial methodological gap by integrating robust statistics with amortized inference. 

The cost of robust ABI was also explored. Compared to a standard estimator, a robust estimator exhibited higher efficiency loss (i.e., higher posterior variance) when applied to uncontaminated data. Nevertheless, the robust neural $\mu$ estimators with Cauchy distribution can be considered as highly robust and efficient compared to its counterparts in traditional robust statistics \citep{hampel_robust_2005, tukey_robust_1979,huber_robust_1964}. Yet, evaluating the efficiency loss of the robust DDM estimator is complicated due to the absence of a direct comparative reference. Additionally, we found the same robustness-efficiency trade-off in ABI as in robust statistics \citep{hampel_robust_2005}: the higher the robustness, the larger the efficiency losses. The cost of robustness in ABI aligned closely with principles from robust statistics, paving the way to integrate these two research domains.

Lastly, we applied the robust ABI to a real data set and found that fitting raw choice reaction time data to the robust estimator with a Cauchy distribution yielded accurate estimates. This method holds significant potential for application in research areas where contamination detection or removal is particularly challenging. 

At the same time, we acknowledge the limitations of current work. In our two examples, we primarily used $t-$based robust estimators, but the choice of contamination distribution is flexible and should be tailored to the specific model and data. This choice requires the researcher to have a general understanding of the potential range of contaminants. Moreover, although the Cauchy distribution is a good contamination distribution candidate when we have univariate data, specifying the contamination distribution in multi-dimensional data is more challenging as it involves accounting for the underlying correlation structure. Thus, it is worthwhile to study this robust approach in a multi-dimensional setting. Secondly, the probability $\pi$ of one entry in the data set being contaminated during network training is determined by the researcher, which again requires knowledge about the empirical data set. A possible extension would involve treating $\pi$ as a parameter that can vary across data sets during data simulation, allowing it to be estimated and used as an index of data quality. Such an approach can also lead to more efficient estimation for data sets that have a lower fraction of outliers. However, estimating $\pi$ could be challenging in cases where the fraction of contaminants is small or when the contamination model differs significantly from the assumed one. Another possible future research direction concerns optimizing the neural architecture or post-training correction \citep{siahkoohi2023reliable} with respect to the robustness of inference. 

In conclusion, as ABI is a powerful method due to its high flexibility and low computational cost in inference, assessing and robustifying the method facilitates both methodological development and practical application. This work paves the way for a new era of robust and efficient Bayesian inference, extending the applicability of ABI to complex real-world data sets.


\bibliography{ref1}

\newpage


\renewcommand\theequation{\Alph{section}\arabic{equation}} 
\counterwithin*{equation}{section} 
\renewcommand\thefigure{\Alph{section}\arabic{figure}} 
\counterwithin*{figure}{section} 
\renewcommand\thetable{\Alph{section}\arabic{table}} 
\counterwithin*{table}{section} 

\newpage

\begin{appendices}

\section*{Appendix}

\subsection*{Parameter Recovery Study for the Drift Diffusion Model}
\setcounter{table}{0}
\renewcommand{\thetable}{A\arabic{table}}
\setcounter{figure}{0}
\renewcommand{\thefigure}{A\arabic{figure}}

\begin{figure}[!ht]
\centering
\includegraphics[width=0.99\linewidth]{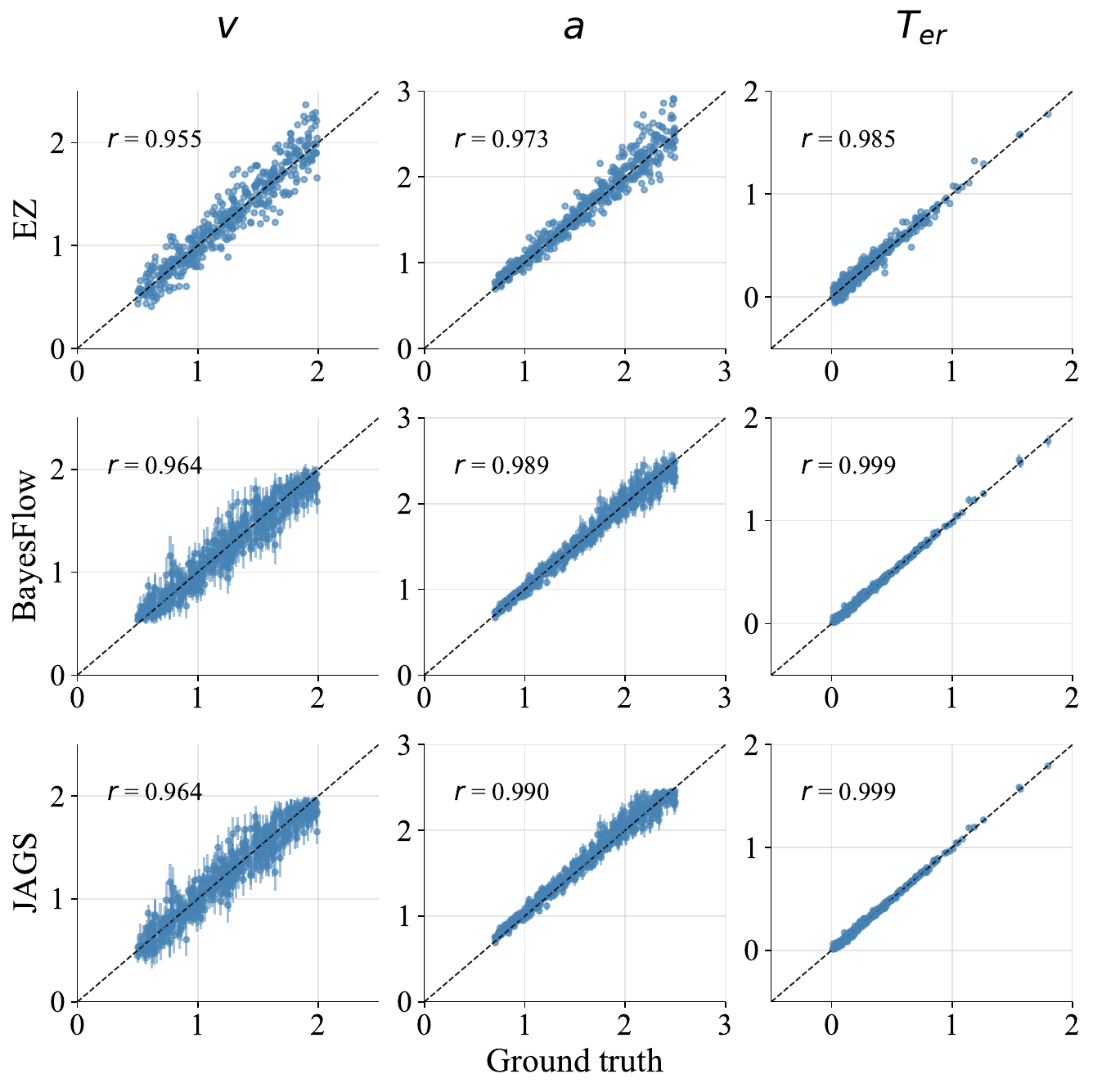}
\caption{\label{fig:prDDM} Parameter recovery of the minimal DDM with different estimators. The x- and y-axis show the ground truth and estimated values, respectively. The diagonal dashed lines indicate perfect recovery. Additionally, the Pearson's correlation coefficient $r$ quantifies the alignment between the estimated and true values. The error bars for JAGS and BayesFlow show $\pm$ posterior standard deviation around the point estimates.}
\end{figure}

\begin{figure}[!ht]
\centering
\includegraphics[width=1\linewidth]{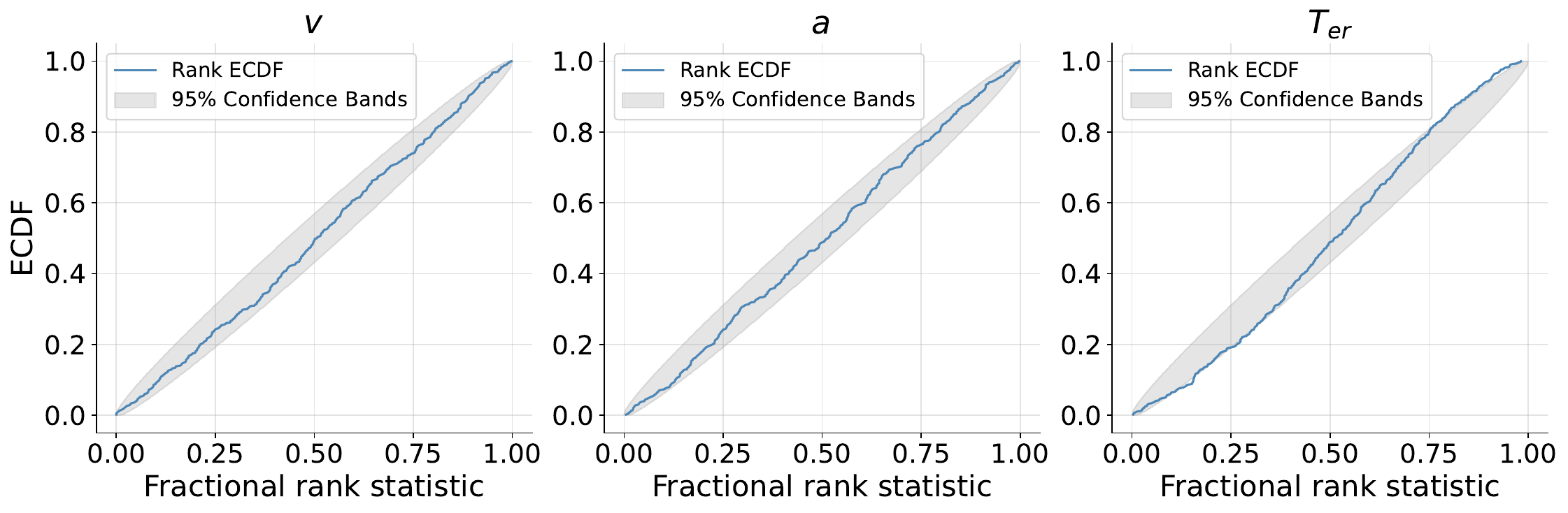 }
\caption{\label{fig:ecdf}The empirical cumulative density function (ECDF) of rank statistics plotted against uniform ECDF. The posterior ranks of the prior draws distributed uniformly, indicating the NPE yielded correct posteriors \citep[for details of rank statistics, please refer to][]{talts_validating_2020}.}
\end{figure}

\begin{figure}[!ht]
\centering
\includegraphics[width=1\linewidth]{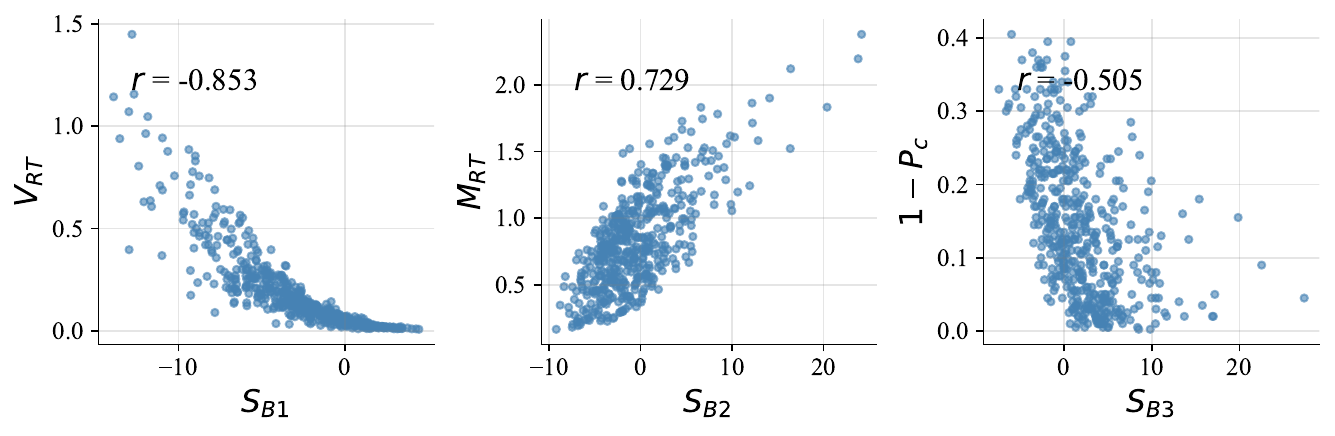 }
\caption{\label{fig:ssvs}The non-linear relationship between $\boldsymbol{s}_B$ and $\boldsymbol{s}_{EZ}$. Three summary statistics from EZ diffusion estimation are plotted against three summary statistics learned by BayesFlow, respectively.}
\end{figure}
\newpage
\subsection*{Priors and Parameter Recoveries in Neural DDM estimators}

\begin{table}[!ht]
\centering
\begin{tabular}{lc}
\toprule
Parameters & Distribution \\
\midrule
$v_{1}$ (Drift Rate 1) & $U(0,7)$  \\
$v_{2}$ (Drift Rate 2) & $U(-7,0)$  \\
$a$ (Boundary Separation) & $U(0.1,5)$  \\
$z$ (Starting Point) & $U(0.01,0.99)$  \\
$T_{er}$ (Non-Decision Time) & $\text{Gamma}(1.5,0.2)$  \\
\bottomrule
\end{tabular}
\captionsetup{} 
\caption{Prior distributions over the DDM parameters. $\mbox{Gamma}(1.5,0.2)$ denotes a gamma distribution with shape parameter 1.5 and scale parameter 0.2.}\label{priorddm}
\end{table}

\begin{figure}[!ht]
\centering
\includegraphics[width=1\linewidth]{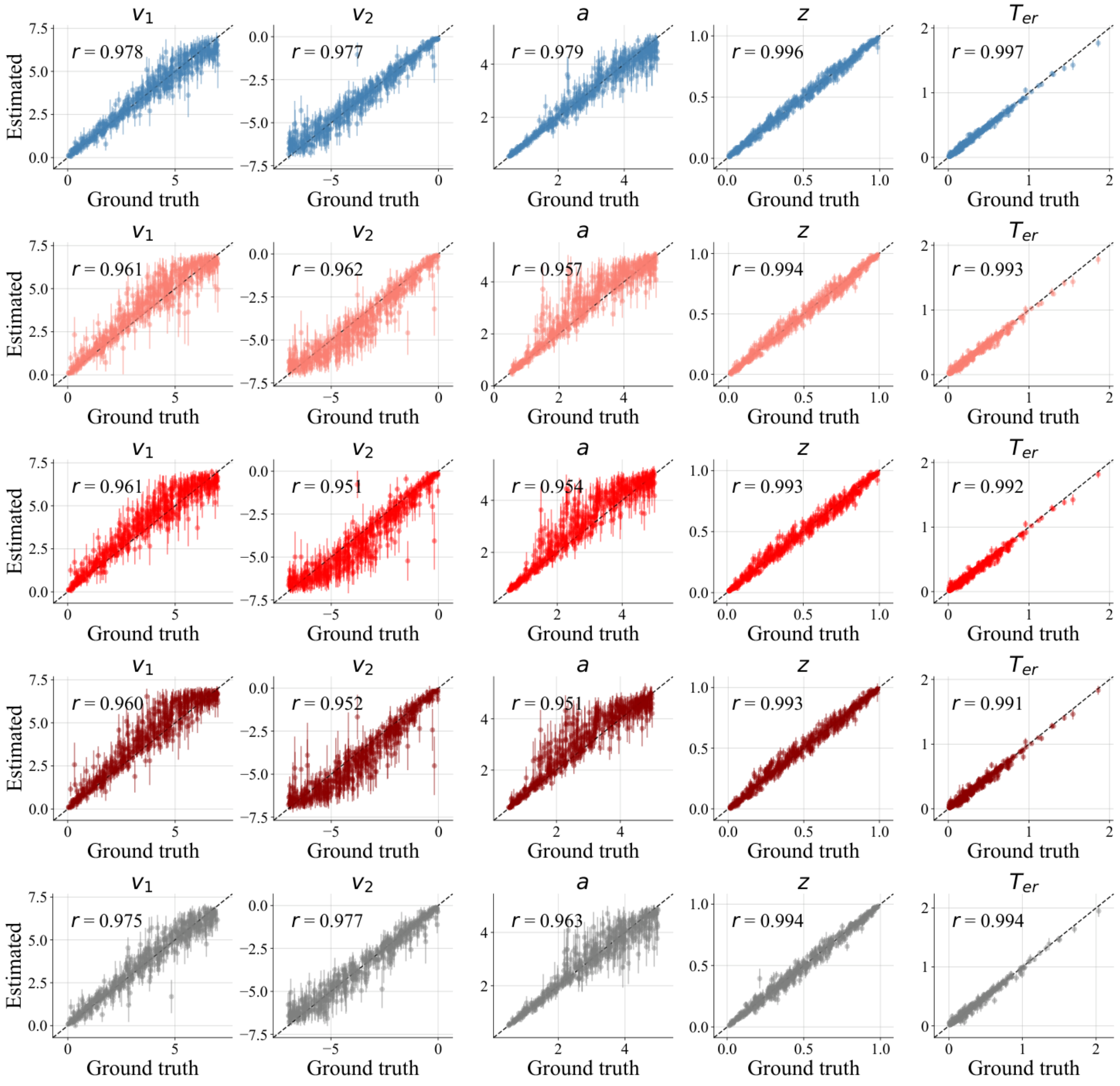 }
\caption{Parameter recovery of DDM estimators on 500 new data sets when the total trial number $n=200$ (100 per condition). The 500 new data set were simulated from the standard DDM simulator without any contamination. From top to bottom, the estimators are: standard estimator, robust estimators with folded-$t_1$, folded-$t_3$, folded-$t_5$, and $U(0, 20)$, respectively.}
\label{fig:prddmrobust}
\end{figure}

\newpage

\subsection*{Robust Estimators with Different Contamination Probability $\pi$}

Through the simulations presented in the main body of this paper, we found $t_1$ distribution the optimal contamination distribution for training robust NPEs, with the contamination probability $\pi=0.1$. We further explored the impact of $\pi$ in both the toy example and the DDM. Specifically, we trained robust NPEs with $\pi \in \{0.01, 0.05, 0.10, 0.20\}$, respectively. 

\begin{figure}[t]
\centering
\includegraphics[width=1\linewidth]{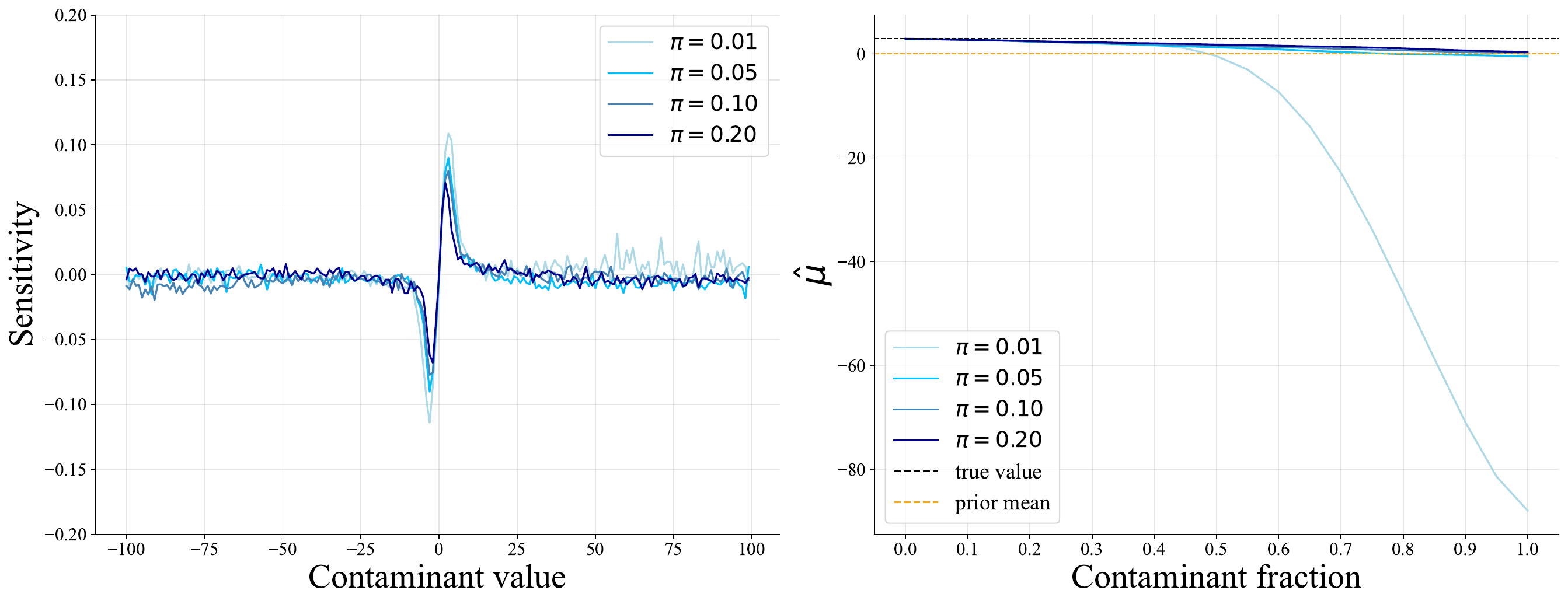 }
\caption{Performance of the $t_1$-based robust $\mu$ estimators with different contamination probability $\pi$. The left panel shows the SSC of the robust estimators with contaminants ranging from -100 to 100 (increasing by 1). The right plot shows the breakdown point with the contaminant value fixed at -100.}
\label{fig:robustnormalpi}
\end{figure}

The network architectures were identical for all the $\mu$ estimators as specified in the main body text. Afterwards, we investigated the robustness of $\mu$ estimators with SSC and BP. As the left panel in Figure~\ref{fig:robustnormalpi} shows, the SSCs show similar shapes for all the robust estimators. All of them were able to filter large contaminants out, and the larger the $\pi$, the lower the sensitivity.  The BP plot (the right panel in Figure~\ref{fig:robustnormalpi}) reveals that the estimator with $\pi=0.01$ had a lower BP around 40\%, whereas the other robust estimators did not break down. Table~\ref{tradeoffnormal} shows us that the MAE ratio and posterior variance ratio of robust estimators decreased with the increase of $\pi$, indicating that the accuracy and efficiency losses were lower for less robust estimators. 

\begin{figure}
\centering
\includegraphics[width=1\linewidth]{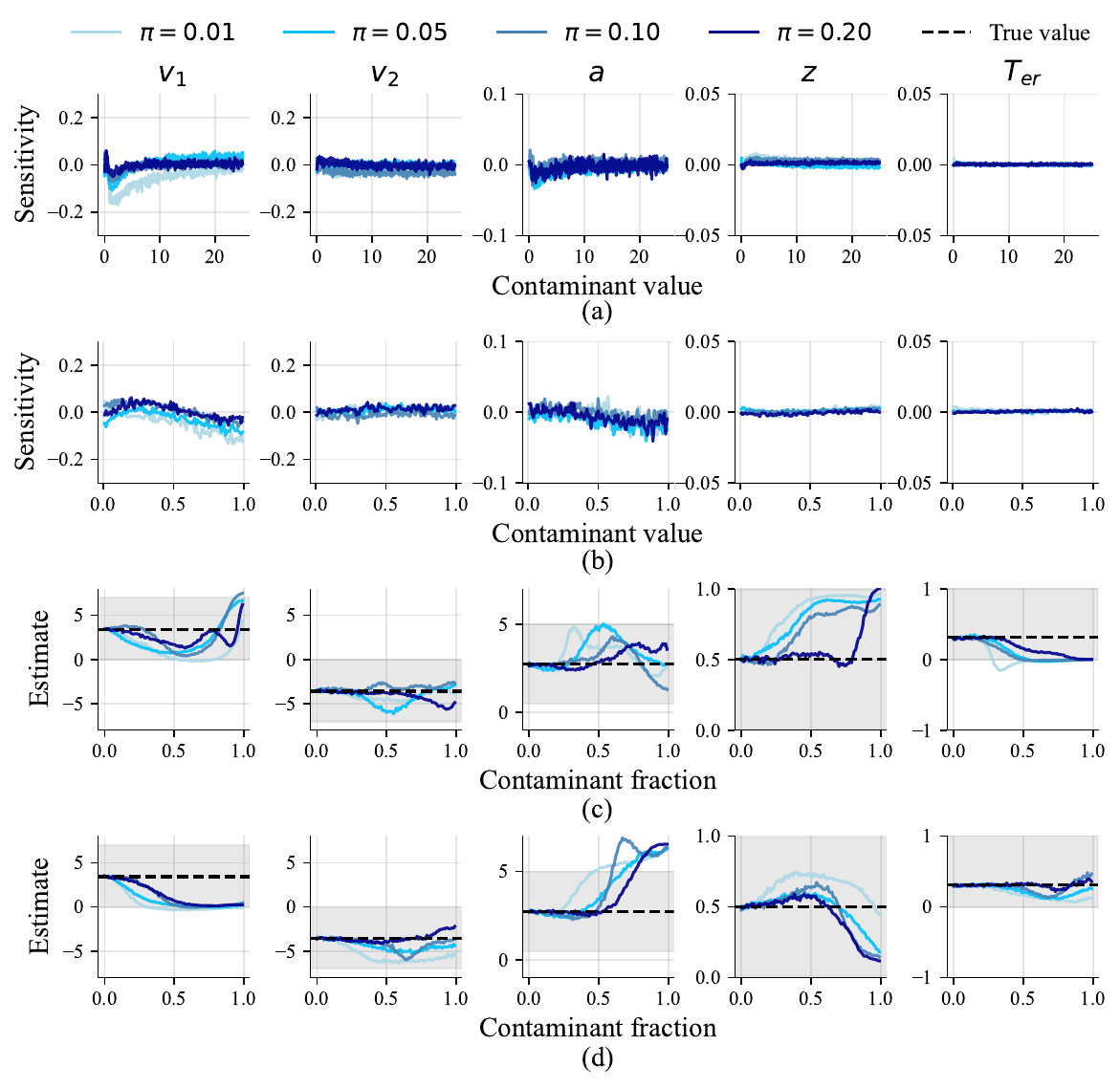 }
\caption{The DDM robust estimators with contamination distribution folded-$t_1$ and different contamination probability $\pi$. Row (a) shows the SSC of the robust estimators with outliers ranging from 0.01s to 25s (increasing by 0.05s). Row (b) shows the SSC with outliers ranging from 0.01s to 1s (increasing by 0.01s). Rows (c) and (d) display the BP plots with outlier values fixed at 0.01s and 20s, respectively.}
\label{fig:robustpi}
\end{figure}

The network architectures and training settings were the same for all the DDM estimators as specified in the main body text. We investigated the SSC and BP of four robust DDM estimators and found that the higher the $\pi$, the less sensitive the estimator to a contaminant, especially when there was a short contaminant (see rows (a) and (b) in Figure~\ref{fig:robustpi}). In terms of BP, the estimators with $\pi = 0.10$ and $0.20$ showed a higher BP than the other two estimators for short contaminants ($x^c = 0.01s$), while the estimators with $\pi = 0.05$ and $0.10$ showed a higher BP for long contaminants ($x^c = 20s$). The robustness and efficiency losses trade-off appeared again in the variance ratio of robust DDM estimators and standard DDM estimators: the higher the $\pi$, the smaller the variance ratio, thus larger efficiency losses. 

To summarize, varying $\pi$ affected the performance of robust estimators. While $\pi=0.01$ can be too low and $\pi=0.20$ is too high in terms of efficiency loss, a $\pi$ from 0.05 to 0.10 allows us to balance these two properties, thus is more suitable to be used in practice.

\begin{table}[!ht]
\centering
\begin{tabular}{lcc}
\toprule
Robust estimator & $\text{MAE Ratio}$ & $\text{Posterior Variance Ratio}$ \\
\midrule
$\pi=0.01$ & 1.003 & 0.981\\
$\pi=0.05$ & 1.000 & 0.945\\
$\pi=0.10$ & 0.996 & 0.913\\
$\pi=0.20$ & 0.978 & 0.863\\
\bottomrule
\end{tabular}
\caption{MAE ratio and posterior variance ratio of location parameter estimators with different contamination probabilities $\pi$. This table shows the MAE ratio and the posterior variance ratio between a standard estimator and the robust estimator. In each row, the robust estimator has a $t_1$ as the contamination distribution and contamination probability $\pi$.}\label{tradeoffnormal}
\end{table}

\newpage

\begin{table}[!ht]
\begin{threeparttable}
\begin{tabular*}{\columnwidth}{@{\extracolsep{\fill}}lccccc@{\extracolsep{\fill}}}
\toprule
Robust estimator & $v_1$ &  $v_2$ & $a$ & $z$ & $T_{er}$ \\
\midrule
$\pi=0.01$ & 0.759 & 0.802 & 0.777 & 0.907 & 0.826 \\
$\pi=0.05$ & 0.73  & 0.732 & 0.735 & 0.892 & 0.797 \\
$\pi=0.10$ & 0.728 & 0.751 & 0.741 & 0.889 & 0.764 \\
$\pi=0.20$ & 0.875 & 0.907 & 0.916 & 0.965 & 0.884 \\
\bottomrule
\end{tabular*}
\caption{MAE ratio in DDM estimators with different $\pi$. This table shows the MAE ratio between a standard estimator and a robust estimator. In each row, the robust estimator has a $\text{folded-}t_1$ as the contamination distribution and a certain contamination probability $\pi$. \label{tradeoffMAE}}
\end{threeparttable}
\end{table}

\begin{table}[!ht]
\begin{threeparttable}
\begin{tabular*}{\columnwidth}{@{\extracolsep\fill}lccccc@{\extracolsep\fill}}
\toprule
Robust estimator & $v_1$ &  $v_2$ & $a$ & $z$ & $T_{er}$  \\
\midrule
$\pi=0.01$ & 0.823 & 0.807 & 0.885 & 0.978 & 0.864 \\
$\pi=0.05$ & 0.720 & 0.699 & 0.74  & 0.819 & 0.71  \\
$\pi=0.10$ & 0.617 & 0.596 & 0.615 & 0.734 & 0.574 \\
$\pi=0.20$ & 0.436 & 0.517 & 0.531 & 0.574 & 0.492 \\
\bottomrule
\end{tabular*}
\caption{Posterior variance ratio of DDM estimators with different $\pi$. This table shows the posterior variance ratio in estimates between a standard estimator and a robust estimator. In each row, the robust estimator has a $\text{folded-}t_1$ as the contamination distribution and a certain contamination probability $\pi$.\label{tradeoffVAR}}
\end{threeparttable}
\end{table}

\newpage
\subsection*{Descriptive Statistics in the Real Data Example}

\begin{figure}[!ht]
\centering
\includegraphics[width=1\linewidth]{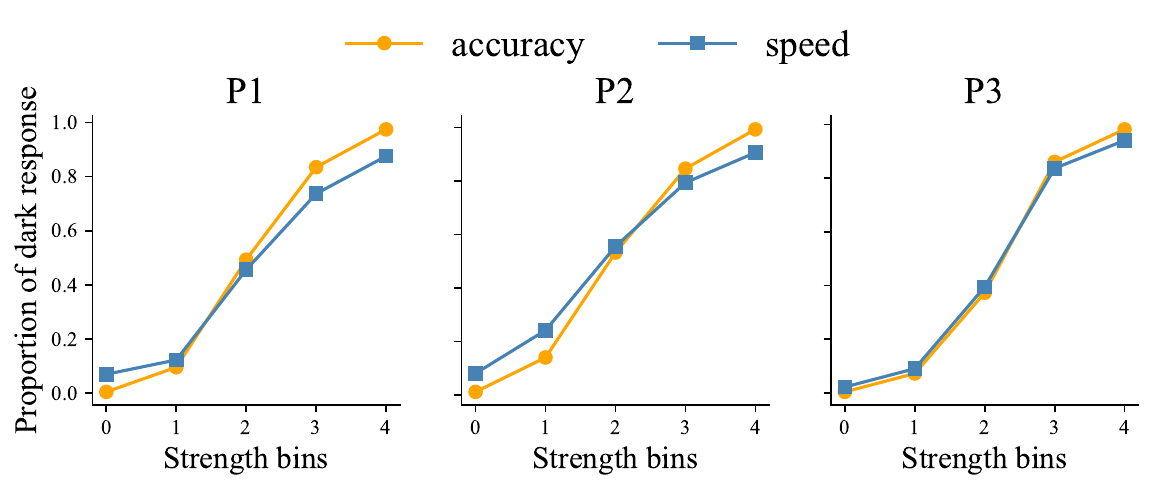}
\caption{\label{fig:ds} The relationship between proportion of ``dark'' responses and the brightness strength bins.}
\end{figure}

\begin{figure}[!ht]
\centering
\includegraphics[width=1\linewidth]{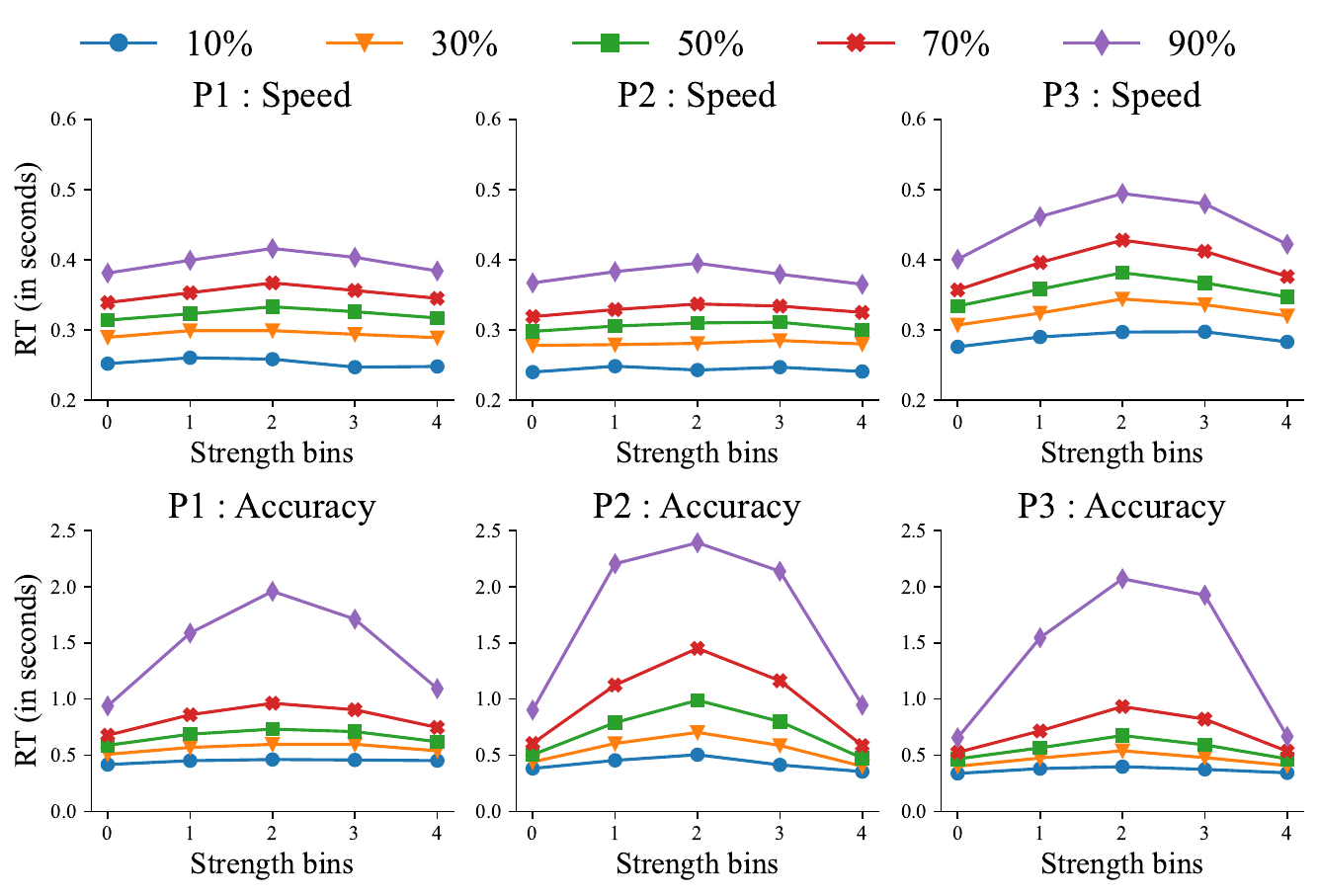}
\caption{Quantile plot of reaction time. The upper panel shows the reaction time quantiles when participants were instructed to respond as fast as possible, while the lower panel displays the reaction time quantiles when participants were instructed to respond as accurately as possible.}
\label{fig:quantile}
\end{figure}
 
\end{appendices}

\end{document}